# Fast Set Bounds Propagation Using a BDD-SAT Hybrid

**Graeme Gange**                                    GGANGE@CSSE.UNIMELB.EDU.AU
**Peter J. Stuckey**                                 PJS@CSSE.UNIMELB.EDU.AU
*National ICT Australia, Victoria Laboratory*
*Department of Computer Science and Software Engineering*
*The University of Melbourne, Vic. 3010, Australia*
**Vitaly Lagoon**                                    LAGOON@CADENCE.COM
*Cadence Design Systems*
*270 Billerica Rd, Chelmsford, MA 01824, USA*

## Abstract

Binary Decision Diagram (BDD) based set bounds propagation is a powerful approach to solving set-constraint satisfaction problems. However, prior BDD based techniques incur the significant overhead of constructing and manipulating graphs during search. We present a set-constraint solver which combines BDD-based set-bounds propagators with the learning abilities of a modern SAT solver. Together with a number of improvements beyond the basic algorithm, this solver is highly competitive with existing propagation based set constraint solvers.

## 1. Introduction

It is often convenient to model a constraint satisfaction problem (CSP) using finite set variables and set relationships between them. A common approach to solving finite domain CSPs is using a combination of backtracking search and a constraint propagation algorithm. The propagation algorithm attempts to enforce consistency on the values in the domains of the constraint variables by removing values from the domains of variables that cannot form part of a complete solution to the system of constraints. The most common level of consistency is *set bounds consistency* (Gervet, 1997) where the solver keeps track for each set of which elements are definitely in or out of the set. Many solvers use set bounds consistency including ECLiPSe (IC-PARC, 2003), GECODE (GECODE, 2008), and ILOG SOLVER (ILOG, 2004).

Set bounds propagation is supported by solvers since stronger notions of propagation such as domain propagation require representing exponentially large domains of possible values. However, Lagoon and Stuckey (2004) demonstrated that it is possible to use reduced ordered binary decision diagrams (BDDs) as a compact representation of both set domains and of set constraints, thus permitting *set domain* propagation. A domain propagator ensures that every value in the domain of a set variable can be extended to a complete assignment of all of the variables in a constraint. The use of the BDD representation comes with several additional benefits. The ability to easily conjoin and existentially quantify BDDs allows the removal of intermediate variables, thus strengthening propagation, and also makes the construction of propagators for global constraints straightforward.

Given the natural way in which BDDs can be used to model set constraint problems, it is therefore worthwhile utilising BDDs to construct other types of set solver. Indeed it has been previously demonstrated (Hawkins, Lagoon, & Stuckey, 2004, 2005) that set bounds propagation can be efficiently implemented using BDDs to represent constraints and domains of variables. A major benefit of the BDD-based approach is that it frees us from the need to laboriously construct set bounds propagators for each new constraint by hand. Moreover, correctness and optimality of such BDD-based propagators follow by construction. The other advantages of the BDD-based representation identified above still apply, and the resulting solver performs very favourably when compared with existing set bounds solvers.

But set bounds propagation using BDDs still constructs BDDs during propagation, which is a considerable overhead. In this paper we show how we can perform BDD-based set bounds propagation using a marking algorithm that perform linear scans of the BDD representation of the constraint without constructing new BDDs. The resulting set bounds propagators are substantially faster than those using BDDs.

The contributions of this paper are:

- **Efficient set bounds propagators**: No new BDDs are constructed during propagation, so it is very fast.

- **Graph reuse**: We can reuse a single BDD for multiple copies of the same constraint, and hence handle larger problems.

- **Ordering flexibility**: We are not restricted to a single global ordering of Booleans for constructing BDDs.

- **Filtering**: We can keep track of which parts of the set variable can really make a difference, and reduce the amount of propagation.

Pure set-bounds propagation tends to perform badly, however, in problems where a large number of similar regions of the search space must be explored. We therefore embed the set-bounds propagators in MiniSAT (Eén & Sörensson, 2003), to provide SAT-style clause learning.

In the next section, we introduce propagation-based solving for set problems, and briefly discuss SAT solving. In Section 3 we discuss binary decision diagrams (BDDs) and how to implement set bounds propagation using BDDs. Then in Section 4, we present the propagation algorithm used by the hybrid solver, together with a number of variations upon the standard algorithm. In Section 5, we show how to incorporate reason generation with BDD propagation to build a hybrid solver In Section 6 we test the performance of the solver on a variety of set-constraint problems, and compare with other set-constraint solvers. In Section 7 we discuss related work, before concluding in Section 8.

## 2. Propagation-based Solving

Propagation based approaches to solving set constraint problems represent the problem using a domain storing the possible values of each set variable, and propagators for each constraint, that remove values from the domain of a variable that are inconsistent with values for other variables. Propagation is combined with backtracking search to find solutions.

A *domain* $D$ is a complete mapping from the fixed finite set of variables $\mathcal{V}$ to finite collections of finite sets of integers. The *domain of a variable* $v$ is the set $D(v)$. A domain $D_1$ is said to be *stronger* than a domain $D_2$, written $D_1 \sqsubseteq D_2$, if $D_1(v) \subseteq D_2(v)$ for all $v \in \mathcal{V}$. A domain $D_1$ is equal to a domain $D_2$, written $D_1 = D_2$, if $D_1(v) = D_2(v)$ for all variables $v \in \mathcal{V}$. A domain $D$ can be interpreted as the constraint $\bigwedge_{v \in \mathcal{V}} v \in D(v)$.

For set constraints we will often be interested in restricting variables to take on *convex* domains. A set of sets $K$ is *convex* if $a, b \in K$ and $a \subseteq c \subseteq b$ implies $c \in K$. We use interval notation $[a, b]$ where $a \subseteq b$ to represent the (minimal) convex set $K$ including $a$ and $b$. For any finite collection of sets $K = \{a_1, a_2, \ldots, a_n\}$, we define the convex closure of $K$: $conv(K) = [\cap_{a \in K} a, \cup_{a \in K} a]$. We extend the concept of convex closure to domains by defining $ran(D)$ to be the domain such that $ran(D)(v) = conv(D(v))$ for all $v \in \mathcal{V}$.

A *valuation* $\theta$ is a set of mappings from the set of variables $\mathcal{V}$ to sets of integer values, written $\{v_1 \mapsto d_1, \ldots, v_n \mapsto d_n\}$. A valuation can be extended to apply to constraints involving the variables in the obvious way. Let *vars* be the function that returns the set of variables appearing in an expression, constraint or valuation. In an abuse of notation, we say a valuation is an element of a domain $D$, written $\theta \in D$, if $\theta(v_i) \in D(v_i)$ for all $v_i \in vars(\theta)$.

## 2.1 Constraints, Propagators and Propagation

A constraint is a restriction placed on the allowable values for a set of variables. We shall use *primitive set constraints* such as (membership) $k \in v$, (equality) $u = v$, (subset) $u \subseteq w$, (union) $u = v \cup w$, (intersection) $u = v \cap w$, (cardinality) $|v| = k$, (upper cardinality bound) $|v| \leq k$, (lexicographic order) $u < v$, where $u, v, w$ are set variables, $k$ is an integer. We can also construct more complicated constraints which are (possibly existentially quantified) conjunctions of primitive set constraints. We define the *solutions* of a constraint $c$ to be the set of valuations $\theta$ on $vars(c)$ that make the constraint true.

We associate a *propagator* with every constraint. A propagator $f$ is a monotonically decreasing function from domains to domains, so $D_1 \sqsubseteq D_2$ implies that $f(D_1) \sqsubseteq f(D_2)$, and $f(D) \sqsubseteq D$. A propagator $f$ is *correct* for a constraint $c$ if and only if for all domains $D$: $\{\theta \mid \theta \in D\} \cap solns(c) = \{\theta \mid \theta \in f(D)\} \cap solns(c)$

A *propagation solver* $solv(F, D)$ for a set of propagators $F$ and a domain $D$ repeatedly applies the propagators in $F$ starting from the domain $D$ until a fixpoint is reached. $solv(F, D)$ is the weakest domain $D' \sqsubseteq D$ where $f(D') = D'$ for all $f \in F$.

**Example 1** A small example of a set-constraint problem would be to, given a universe consisting of the elements $\{1, 2, 3, 4\}$, find values for variables $x, y, z$ such that $z = x \cap y$, $|x| = 3$, $|y| = 3$, $|z| = 2$, $3 \notin z$, $1 \in z$ and $2 \notin y$.

The unique solution to this problem is $\theta = \{x \mapsto \{1, 2, 4\}, y \mapsto \{1, 3, 4\}, z \mapsto \{1, 4\}\}$.

## 2.2 Set Bounds Consistency

A domain $D$ is *(set) bounds consistent* for a constraint $c$ if for every variable $v \in vars(c)$ the upper bound of $D(v)$ is the union of the values of $v$ in all solutions of $c$ in $D$, and the lower bound of $D(v)$ is the intersection of the values of $v$ in all solutions of $c$ in $D$. We define the *set bounds propagator* for a constraint $c$ as

$$ub(c)(D)(v) = \begin{cases} \{i \mid \exists \theta \cdot \theta \in solns(D \wedge c) \wedge i \in \theta(v)\} & \text{if } v \in vars(c) \\ ub(v) & \text{otherwise} \end{cases}$$

$$lb(c)(D)(v) = \begin{cases} \{i \mid \forall \theta \cdot \theta \in solns(D \wedge c) \rightarrow i \in \theta(v)\} & \text{if } v \in vars(c) \\ lb(v) & \text{otherwise} \end{cases}$$

$$sb(c)(D)(v) = [lb(c)(D)(v), ub(c)(D)(v)]$$

Then $sb(c)(D)$ is always bounds consistent with $c$.

**Example 2** Continuing the example from the previous section, the initial bounds of the variables $x, y, z$ are $D(x) = D(y) = D(z) = [\emptyset, \{1, 2, 3, 4\}]$, as no values are explicitly included or excluded from the domains. As first $3 \notin z$ is added, then $1 \in z$ and finally $2 \notin y$, the bounds are reduced, and the consequences of these changes are propagated among the variables as follows:

| Propagator | $D(x)$ | $D(y)$ | $D(z)$ |
|---|---|---|---|
| | $[\emptyset, \{1, 2, 3, 4\}]$ | $[\emptyset, \{1, 2, 3, 4\}]$ | $[\emptyset, \{1, 2, 3, 4\}]$ |
| $3 \notin z$ | $[\emptyset, \{1, 2, 3, 4\}]$ | $[\emptyset, \{1, 2, 3, 4\}]$ | $[\emptyset, \{1, 2, 4\}]$ |
| $1 \in z$ | $[\emptyset, \{1, 2, 3, 4\}]$ | $[\emptyset, \{1, 2, 3, 4\}]$ | $[\{1\}, \{1, 2, 4\}]$ |
| $z = x \cap y$ | $[\{1\}, \{1, 2, 3, 4\}]$ | $[\{1\}, \{1, 2, 3, 4\}]$ | $[\{1\}, \{1, 2, 4\}]$ |
| $2 \notin y$ | $[\{1\}, \{1, 2, 3, 4\}]$ | $[\{1\}, \{1, 3, 4\}]$ | $[\{1\}, \{1, 2, 4\}]$ |
| $|y| = 3$ | $[\{1\}, \{1, 2, 3, 4\}]$ | $[\{1, 3, 4\}, \{1, 3, 4\}]$ | $[\{1\}, \{1, 2, 4\}]$ |
| $z = x \cap y$ | $[\{1\}, \{1, 2, 4\}]$ | $[\{1, 3, 4\}, \{1, 3, 4\}]$ | $[\{1\}, \{1, 4\}]$ |
| $|z| = 2$ | $[\{1\}, \{1, 2, 4\}]$ | $[\{1, 3, 4\}, \{1, 3, 4\}]$ | $[\{1, 4\}, \{1, 4\}]$ |
| $|x| = 3$ | $[\{1, 2, 4\}, \{1, 2, 4\}]$ | $[\{1, 3, 4\}, \{1, 3, 4\}]$ | $[\{1, 4\}, \{1, 4\}]$ |

Once $1 \in z$ is fixed, 1 is added to $lb(z)$. Since $z = x \cap y$, any element in $lb(z)$ must also be in $lb(x)$ and $lb(y)$. Once $2 \notin y$ has been set, $|ub(y)| = 3$ and since $|ub(y)| \geq |y| = 3$ this means $y = ub(y) = \{1, 3, 4\}$. This means that $2 \notin z$ since $z = x \cap y$. Since $3 \notin ub(z)$, at least one of $x$ or $y$ must not contain 3. Once $3 \in lb(y)$ has set, it can be determined that $3 \notin ub(x)$. Since $|ub(z)| = 2$ this forces $z = ub(z) = \{1, 4\}$. Finally the constraint $|x| = 3$ then results in the value of $x$ becoming fixed. The corresponding valuation is $\theta = \{x \mapsto \{1, 2, 4\}, y \mapsto \{1, 3, 4\}, z \mapsto \{1, 4\}\}$, which is the solution provided in Example 1.

## 2.3 Boolean Satisfiability (SAT)

Boolean Satisfiability or SAT solvers are a special case of propagation-based solvers, restricted to Boolean variables and clause constraints.

The Davis-Putnam-Logemann-Loveland (DPLL) algorithm (Davis, Logemann, & Loveland, 1962), on which most modern SAT solvers are based, is also a propagation-based approach to solving SAT problems. It interleaves two phases – search, where an unfixed variable is assigned a value, and propagation (so called unit propagation).

Modern SAT solvers incorporate sophisticated engineering to propagate constraints very fast, to record as nogoods part of the search that lead to failure, and to automate the search by keeping track of how often a variable is part of the reason for causing failure (*activity*) and concentrating search on variables with high activity. Modern SAT solvers also frequently restart the search from scratch relying on nogoods recording to prevent repeated search, and activity to drive the search into more profitable areas. See e.g., the report by Eén and Sörensson (2003) for a good introduction to modern SAT solving.

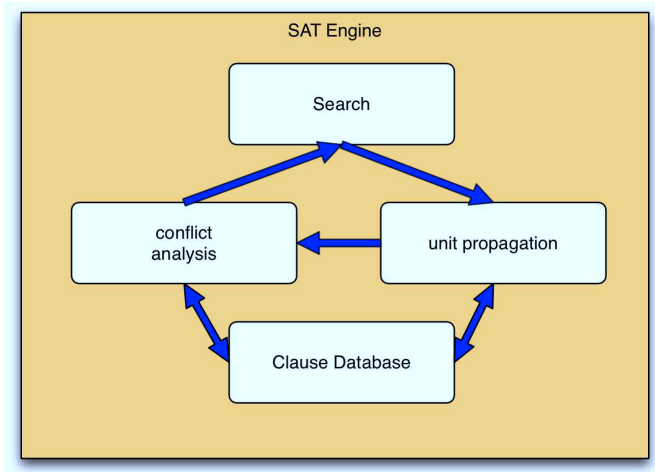

Figure 1: Architecture for the SAT solver.

A rough architecture of a modern SAT solver is illustrated in Figure 1. Search starts the unit propagation process which interacts with the clause database and may detect failure, which initiates conflict analysis. Unit propagation records for each literal that is made true, the clause that explains why the literal become true. Conflict analysis uses the graph of explanations to construct a nogood which is a resolvent of clauses causing the failure that adds to the strength of unit propagation. This is stored in the clause database and causes search to backjump. It prevents the search revisiting the same set of decisions. Not detailed here are activity counters which record which variables are most responsible for failure, these are the variables chosen for labelling by the search.

## 3. Binary Decision Diagrams

We assume a set $\mathcal{B}$ of Boolean variables with a total ordering $\prec$. A Boolean variable can take the value 0 (false) or 1 (true). We make use of the following Boolean operations: $\wedge$ (conjunction), $\vee$ (disjunction), $\neg$ (negation), $\rightarrow$ (implication), $\leftrightarrow$ (bi-implication) and $\exists$ (existential quantification). We denote by $\exists_V F$ the formula $\exists x_1 \cdots \exists x_n F$ where $V = \{x_1, \ldots, x_n\}$, and by $\bar{\exists}_V F$ we mean $\exists_{V'} F$ where $V' = vars(F) \setminus V$.

Reduced Ordered Binary Decision Diagrams are a well-known method of representing Boolean functions on Boolean variables using directed acyclic graphs with a single root. Every internal node $n(v, f, t)$ in a BDD $r$ is labelled with a Boolean variable $v \in \mathcal{B}$, and has two outgoing arcs — the 'false' arc (to BDD $f$) and the 'true' arc (to BDD $t$). Leaf nodes are either $\mathcal{F}$ (false) or $\mathcal{T}$ (true). Each node represents a single test of the labelled variable; when traversing the tree the appropriate arc is followed depending on the value of the variable. Define the size $|r|$ as the number of internal nodes in a BDD $r$, and $VAR(r)$ as the set of variables $v \in \mathcal{B}$ appearing in some internal node in $r$.

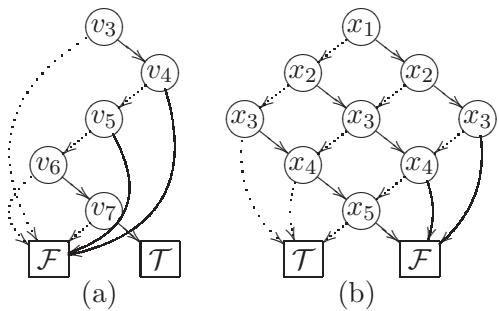

Figure 2: BDDs for (a) $v_3 \wedge \neg v_4 \wedge \neg v_5 \wedge v_6 \wedge v_7$. (b) $x_1 + x_2 + x_3 + x_4 + x_5 \leq 2$. The node $n(v, f, t)$ is shown as a circle labelled $v$ with a dotted arc to the $f$ BDD, and a solid arc to the $t$ BDD.

Reduced Ordered Binary Decision Diagrams (BDDs) (Bryant, 1986) require that the BDD is: *reduced*, that is it contains no identical nodes (nodes with the same variable label and identical true and false arcs) and has no redundant tests (no node has both true and false arcs leading to the same node); and *ordered*, if there is an arc from a node labelled $v_1$ to a node labelled $v_2$ then $v_1 \prec v_2$. A BDD has the nice property that the function representation is canonical up to variable reordering. This permits efficient implementations of many Boolean operations.

A Boolean variable $v$ is said to be *fixed* in a BDD $r$ if either for every node $n(v, f, t) \in r$ $t$ is the constant $\mathcal{F}$ node, or for every node $n(v, f, t)$ $f$ is the constant $\mathcal{F}$ node. Such variables can be identified in a linear time scan over the domain BDD (see e.g., Hawkins et al., 2005). For convenience, if $\phi$ is a BDD, we write $[\![\phi]\!]$ to denote the BDD representing the conjunction of the fixed variables of $\phi$.

**Example 3** Figure 2(a) gives an example of a BDD representing the formula $v_3 \wedge \neg v_4 \wedge \neg v_5 \wedge v_6 \wedge v_7$. Figure 2(b) gives an example of a more complex BDD representing the formula $x_1 + x_2 + x_3 + x_4 + x_5 \leq 2$ where we interpret the Booleans as 0-1 integers. One can verify that the valuation $\{x_1 \mapsto 1, x_2 \mapsto 0, x_3 \mapsto 1, x_4 \mapsto 0, x_5 \mapsto 0\}$ makes the formula true by following the path right, left, right, left, left from the root.

## 3.1 Set Propagation using BDDs

The key step in building set propagation using BDDs is to realize that we can represent a finite set domain using a BDD.

### 3.1.1 Representing domains

If $v$ is a set variable ranging over subsets of $\{1, \ldots, N\}$, then we can represent $v$ using the Boolean variables $V(v) = \{v_1, \ldots, v_N\} \subseteq \mathcal{B}$, where $v_i$ is true iff $i \in v$. We will order the variables $v_1 \prec v_2 \cdots \prec v_N$. We can represent a valuation $\theta$ using a formula

$$R(\theta) = \bigwedge_{v \in vars(\theta)} \left( \bigwedge_{i \in \theta(v)} v_i \wedge \bigwedge_{i \in \{1,\dots,N\} - \theta(v)} \neg v_i \right).$$

Then a domain of variable $v$, $D(v)$ can be represented as $\phi = \bigvee_{a \in D(v)} R(\{v \mapsto a\})$. This formula can be represented by a BDD. The set bounds of $v$ can be obtained by extracting the fixed variables from this BDD, $[\![\phi]\!]$.

For example the valuation $\theta$ of Example 1 is represented by the formula $R(\theta)$:

$$x_1 \wedge x_2 \wedge \neg x_3 \wedge x_4 \wedge y_1 \wedge \neg y_2 \wedge y_3 \wedge y_4 \wedge z_1 \wedge \neg z_2 \wedge \neg z_3 \wedge z_4.$$

And the domain $D(v) = [\{3, 6, 7\}, \{1, 2, 3, 6, 7, 8, 9\}]$ is represented by the BDD in Figure 2(a) since $v_3$, $v_6$ and $v_7$ are true so $3, 6, 7$ are definitely in the set, and $v_4$ and $v_5$ are false so $4$ and $5$ are definitely not in the set.

### 3.1.2 Representing constraints

We can similarly model any set constraint $c$ as a BDD $B(c)$ using the Boolean variable representation $V(v)$ of its set variables $v$. By ordering the variables in each BDD carefully we can build small representations of the formulae. The *pointwise* order of Boolean variables is defined as follows. Given set variables $u \prec v \prec w$ ranging over sets from $\{1, \dots, N\}$ we order the Boolean variables as $u_1 \prec v_1 \prec w_1 \prec u_2 \prec v_2 \prec w_2 \prec \cdots u_N \prec v_N \prec w_N$.

The representation $B(c)$ is simply $\vee_{\theta \in solns(c)} R(\theta)$. For primitive set constraints (using the pointwise order) this size is linear in $N$. For more details see the work of Hawkins et al. (2005). The BDD representation of $|x| \le 2$ is shown in Figure 2(b), for $N = 5$.

### 3.1.3 BDD-based Set Bounds Propagation

We can build a set bounds propagator, more or less from the definition, since we have BDDs to represent domains and constraints.

$$\phi = B(c) \wedge \bigwedge_{v' \in vars(c)} D(v')$$
$$sb(c)(D)(v) = \overline{\exists}_{V(v)} [\![\phi]\!]$$

We simply conjoin the domains to the constraint obtaining $\phi$, then extract the fixed variables from the result, and then project out the relevant part for each variable $v$. The set bounds propagation can be improved by removing the fixed variables as soon as possible. The improved definition is given by Hawkins et al. (2004). Overall the complexity can be made $O(|B(c)|)$.

The updated set bounds can be used to simplify the BDD representing the propagator. Since fixed variables will never interact further with propagation they can be projected out of $B(c)$, so we can replace $B(c)$ by $\exists_{VAR([\![\phi]\!])}\phi$.

```
bdd2sat(node) {
    switch node {
    F: return (0, {}) ;
    T: return (1, {});
    n(v, f, t):
        if (visit[node] ≠ ⊥) return(visit[node],{});
        let n′ be a new Boolean variable;
        visit[node] = n′;
        (f′, C_f) = bdd2sat(f);
        (t′, C_t) = bdd2sat(t);
        return (n′, {v ∧ t′ → n′, ¬v ∧ f′ → n′, v ∧ ¬t′ → ¬n′, ¬v ∧ ¬f′ → ¬n′,
                      t′ ∧ f′ → n′, ¬t′ ∧ ¬f′ → ¬n′} ∪ C_f ∪ C_t);
    }
}
```

Figure 3: Pseudo-code for Tseitin transformation of BDD rooted at $node$ where $n'$ is the Boolean variable encoding the truth value of $node$.

## 3.2 Tseitin Transformation

It is possible to convert any Boolean circuit to a pure SAT representation; the method for doing so is generally attributed to Tseitin (1968). Figure 3 gives pseudo code for the translation of a BDD rooted at $node$, returning a pair of (Boolean variable, set of clauses). The clauses enforce that the Boolean variable takes the truth value of the BDD. Like most BDD algorithms it relies on marking the visited nodes to ensure each node is visited at most once. It assumes the array $visit[]$ is initially all bottom $\bot$, and on first visiting a node stores the corresponding Boolean variable in $visit[]$. A more comprehensive discussion of the Tseitin transformation is presented by Eén and Sörensson (2006).

The constraint is enforced by fixing the variable corresponding to the root node to $true$. An advantage of replacing a BDD by its Tseitin representation is that we can use an unmodified SAT solver to then tackle BDD-based set constraint problems. We shall see in Section 6 that this approach cannot compete with handling the BDDs directly.

## 4. Faster Set-bounds Propagation

While set bounds propagation using BDDs is much faster than set domain propagation and often better than set domain propagation (or other variations of propagation for sets) it still creates new BDDs. This is not necessary as long as we are prepared to give up the simplifying of BDDs that is possible in set bounds propagation.

We do not represent domains of variables as BDDs, but rather as arrays of Boolean domains. A domain $D$ is an array where, for variable $v$ ranging over subsets of $\{1, \ldots, N\}$: $0 \notin D[v_i]$ indicates $i \in v$, and $1 \notin D[v_i]$ indicates $i \notin v$. If $D[v_i] = \{0, 1\}$, we don't know whether $i$ is in or not in $v$. Hence $D(v) = [\{i | 0 \notin D[v_i]\}, \{i | 1 \in D[v_i]\}]$.

The BDD representation of a constraint $B(c)$ is built as before. A significant difference is that since constraints only communicate through the set bounds of variables we do not need them to share a global variable order hence we can if necessary modify the variable order used to construct $B(c)$ for each $c$, or use automatic variable reordering (which is available in most BDD packages) to construct $B(c)$. Another advantage is that we can reuse the BDD for a constraint $c(\bar{x})$ on variables $\bar{x}$ for the constraint $c(\bar{y})$ on variables $\bar{y}$ (as long as they range over the same initial sets), that is, the same constraint on different variables. Hence we only have to build one such BDD, rather than one for each instance of the constraint.

The set bounds propagator $sb(c(\bar{x}))$ for constraint $c(\bar{x})$ is now implemented as follows. A generic BDD representation $r$ of the constraint $c(\bar{y})$ is constructed. The propagator copies the domain description of the actual parameters $x^1, \ldots, x^n$ onto a domain description $E$ for formal parameters $y^1, \ldots, y^n$. It constructs an array $E$ where $E[y_i^j] = D[x_i^j]$. Let $V = \{y_i^j \mid 1 \leq j \leq n, 1 \leq i \leq N\}$ be the set of Boolean variables occurring in the constraint $c(\bar{y})$. The propagator executes the code $\mathsf{bddprop}(r, V, E)$ shown in Figures 4 and 5 which returns $(r', V', E')$. If $r' = \mathcal{F}$ the propagator returns a false domain, otherwise the propagator copies back the domains of the formal parameters to the actual parameters so $D[x_i^j] = E[y_i^j]$. We will come back to the $V'$ argument in the next subsection.

The procedure $\mathsf{bddprop}(r, V, E)$ traverses the BDD $r$ as follows. We visit each node $n(v, f, t)$ in the BDD in a top-down memoing manner. We record if, under the current domain, the node can reach the $\mathcal{F}$ node, and if it can reach the $\mathcal{T}$ node. If the $f$ child can reach the $\mathcal{T}$ node we add support for the variable $v$ taking value 0. Similarly if the $t$ child can reach $\mathcal{T}$ we add support for the variable $v$ taking value 1. If the node can reach both $\mathcal{F}$ and $\mathcal{T}$ we record that the variable $v$ matters to the computation of the BDD. After the visit we reduce the variable set for the propagator to those that matter, and remove values with no support from the domain. The procedure assumes a global *time* variable which is incremented between each propagation, which is used to memo the marking phase. The $top(n, V)$ function returns the variable in the root node of $n$ or the largest variable (under $\prec$) in $V$ if $n = \mathcal{T}$ or $n = \mathcal{F}$.

As presented $\mathsf{bddprop}$ has time complexity $O(|r| \times |V|)$ where $|r|$ is the number of nodes appearing in BDD $r$. In practice the complexity is $O(|r| + |V|)$ since the $|V|$ factor arises from handling "long arcs", where a node $n(v, f, t)$ has a child node ($f$ or $t$) are labelled by a Boolean different from that next in the order $\prec$ after $v$. For set constraints the length of a long arc is typically bounded by the arity of the set constraint. It is possible to create a version of $\mathsf{bddprop}$ which is strictly $O(|r|)$ by careful handling of long arcs. We did so, but in practice it was slower than the form presented here. $\mathsf{bddprop}$ has space complexity $O(|V| + |r|)$ the first component for maintaining the domains of variables and the second for memoing the BDD nodes.

**Example 4** Consider the BDD for the constraint $x = y \cup z$ when $N = 2$ shown in Figure 6(a). Assuming a domain $E$ where $E[y_1] = \{1\}$ ($1 \in y$) and $E[z_2] = \{1\}$ ($2 \in z$), and the remaining variables take value $\{0, 1\}$, the algorithm traverses the edges shown with double lines in Figure 6(b). No path from $x_1$, or $x_2$ following the $f$ arc reaches $\mathcal{T}$ hence 0 is not added to $E'[x_1]$ or $E'[x_2]$. As a result $E[x_1]$ and $E[x_2]$ are set to $\{1\}$. Hence we have determined $1 \in x$ and $2 \in x$.

Also, no nodes for $z_1$ are actually visited, and the left node for $y_2$ only reaches $\mathcal{F}$ and the right node only reaches $\mathcal{T}$. Hence $matters[z_1]$ and $matters[y_2]$ are not marked with the

```
bddprop(r,V,E) {
    for (v ∈ V) {
        E′[v] = {};
    }
    (reachf, reacht) = bddp(r, V, E);
    if (¬reacht) return (F, ∅, E);
    vars = ∅;
    for (v ∈ V) {
        if (E′[v] ≠ E[v]) {
            E[v] = E′[v];
        }
        if (E[v] = {0, 1} ∧ matters[v] ≥ time) vars = vars ∪ {v};
    }
    return (r, vars, E);
}
```

Figure 4: Pseudo-code for BDD-propagation.

current time. The set of *vars* collected by bddprop is empty, since the remaining variables are fixed.

## 4.1 Waking up Less Often

In practice a bounds propagation solver does not blindly apply each propagator until fixpoint, but keeps track of which propagators must still be at fixpoint, and only executes those that may not be. For set bounds this is usually managed as follows. To each set variable $v$ is attached a list of propagators $c$ that involve $v$. Whenever $v$ changes, these propagators are rescheduled for execution.

We can do better than this with the BDD based propagators. The algorithm bddprop collects the set of Boolean variables that matter to the BDD, that is can change the result. If a variable that does not matter becomes fixed, then set bounds propagation cannot learn any new information. We modify the wakeup process as follows. Each propagator stores a list *vars* of Boolean variables which matter given the current domain. When a Boolean variable $x_i^j$ becomes fixed we traverse the list of propagators involving $x_i^j$ and wake those propagators where $x_i^j$ occurs in *vars*. On executing a propagator we revise the set *vars* stored in the propagator. Note the same optimization could be applied to the standard approach, but requires the overhead of computing *vars* which here is folded into bddprop. It is possible to instead do propagator wake-up on literals, rather than variables. In this case, we observe that fixing a variable $v$ to true matters to a node $n(v, f, t)$ iff $\mathcal{T}$ is reachable from $f$ and $\mathcal{F}$ is reachable from $t$ – the converse holds for $\neg v$. In terms of the pseudo-code in Figure 5, the line

**if** $(reachf \wedge reacht)$ $matters[v] = time;$

may therefore be replaced with

```
bddp(node,V,E) {                          ← if (in_set(fset, node)) { return (1,0)};
    switch node {
    F: return (1,0);
    T: return (0,1);
    n(v, f, t):
        if (visit[node] ≥ time) return save[node];
        reachf = 0; reacht = 0;
        if (0 ∈ E[v]) {
            (rf0, rt0) = bddp(f, V, E);
            reachf = reachf ∨ rf0;
            reacht = reacht ∨ rt0;
            if (rt0) {
                for (v' ∈ V, v ≺ v' ≺ top(f, V))
                    E'[v'] = E[v'];
                E'[v] = E'[v] ∪ 0;
            }
        }
        if (1 ∈ E[v]) {
            (rf1, rt1) = bddp(t, V, E);
            reachf = reachf ∨ rf1;
            reacht = reacht ∨ rt1;
            if (rt1) {
                for (v' ∈ V, v ≺ v' ≺ top(t, V))
                    E'[v'] = E[v'];
                E'[v] = E'[v] ∪ 1;
            }
        }
        if (reachf ∧ reacht) matters[v] = time;
        save[node] = (reachf, reacht);   ← if (¬reacht) { insert(fset, node) };
        visit[node] = time;
        return (reachf, reacht);
    }
}
```

Figure 5: Pseudo-code for processing the constraint graph during propagation. Modifications necessary for using dead-subgraph memoization are shown on the right.

```
        if (rt0 ∧ rf1) matters[v] = time;
        if (rt1 ∧ rf0) matters[¬v] = time;
```

While this allows for propagators to wake up less frequently, propagator execution is slower due to keeping track of additional reachability information.

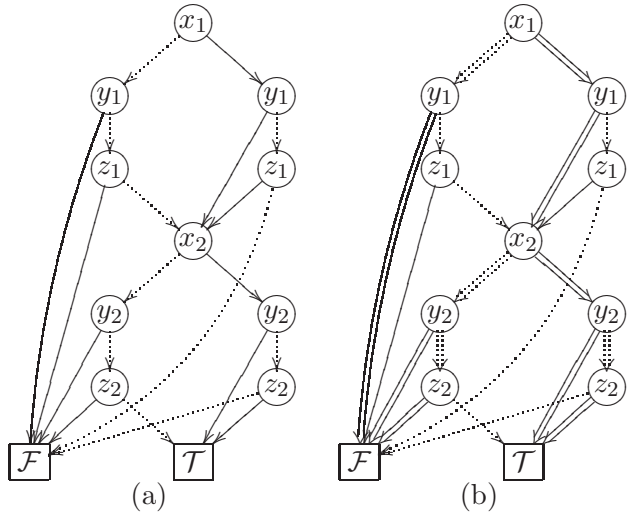

Figure 6: (a) The BDD representing $x = y \cup z$ where $N = 2$. (b) The edges traversed by bddprop, when $E[y_1] = \{1\}$ and $E[z_2] = \{1\}$ and $E[v] = \{0, 1\}$ otherwise, are shown doubled.

## 4.2 Dead Subgraph Memoization and Shortcutting

The algorithm as presented above always explores all reachable parts of the graph in order to determine the set of supported values. However, a number of improvements for Multi-Decision Diagrams (MDDs) were presented by Cheng and Yap (2008) which reduce the portion of the graph which must be traversed in order to enforce consistency. These are dead subgraph memoization, which avoids traversal of subgraphs that cannot provide support for any values, and shortcutting, which recognizes situations where it is only necessary to find one path to $\mathcal{T}$ to ensure consistency. These can readily be adapted to a BDD-based set constraint solver.

### 4.2.1 DEAD SUBGRAPH MEMOIZATION

The key observation for dead subgraph memoization is that, as search progresses, paths along the graph to $\mathcal{T}$ are only ever removed. As such, if $\mathcal{T}$ becomes unreachable from a node $n$, the subgraph incident from $n$ need never again be explored until the solver backtracks. Thus, if the set of dead nodes can be maintained, it is possible to progressively eliminate subgraphs during propagation.

We keep for each instance of a constraint $c(\bar{x})$ a *failure set*, *fset* which records which nodes can not reach $\mathcal{T}$ (and hence are equivalent to $\mathcal{F}$). During propagation, once a node $n$ is shown to have no path to $\mathcal{T}$, it is added to the failure set *fset*. When a node is processed, we first check if it is in *fset*—if so, we terminate early, otherwise we proceed as normal. The modifications necessary for this are shown on the right in Figure 5. For simplicity the pseudo-code treats *fset* as a global.

A method for efficiently maintaining the failure sets was presented by Cheng and Yap (2008), which uses sparse-set data structures to provide efficient lookup, insertion and backtracking. The set *fset* is maintained as a pair of arrays: *sparse* and *dense* and a counter

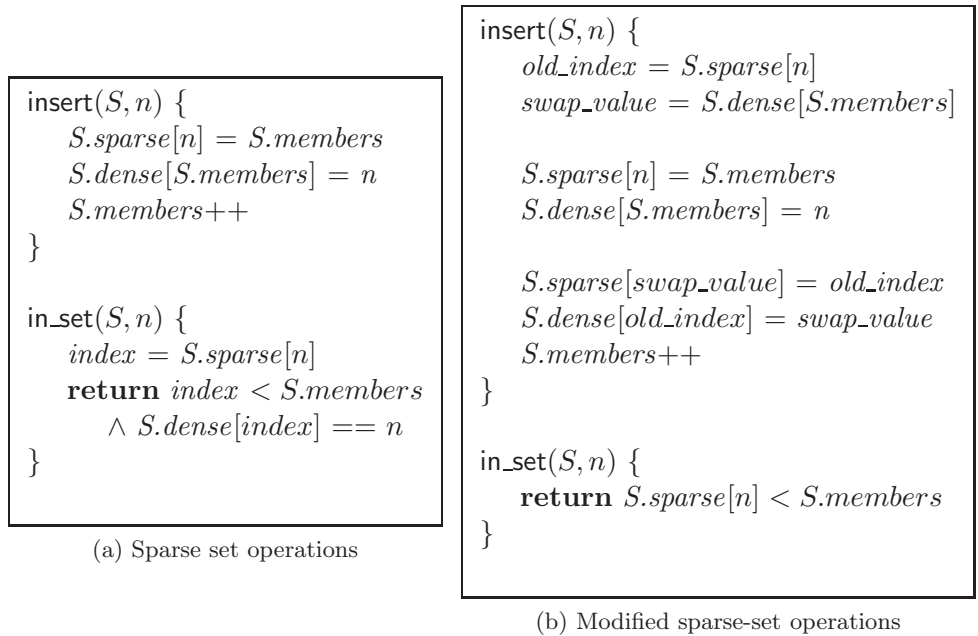

(a) Sparse set operations

(b) Modified sparse-set operations

Figure 7: Pseudo-code for conventional sparse-set operations, and the corresponding modified versions.

*members*. $n \in fset$ if $sparse[n] < members$ and $dense[sparse[n]] = n$. The operations for insertion and testing are shown in Figure 7(a). Crucially we can backtrack to earlier forms of the set simply by resetting *members* to its value at that time.

These structures can be improved slightly by the observation that checking membership will occur significantly more often than insertion. Pseudo-code for the modified sparse-set operations are given in Figure 7(b). While insertion operations become more expensive, the overall computation time is reduced.

**Example 5** Consider the set illustrated in Figure 8(a). The elements in the set are $\{1, 7\}$. We can determine that the element 4 is not in the set $S_0$, as $sparse[4]$ is not strictly less than *members*, indicated by the arrow in Figure 8(a).

To insert an element $v$ using the standard sparse-set operations, we merely overwrite $dense[members]$ with $v$, and set the value of $sparse[v]$ to *members*. This is shown in Figure 8(b), inserting 3 into $S_0$. At this point, both $sparse[3]$ and $sparse[4]$ have the value 2. To test if $4 \in S'_0$, it is not sufficient to determine if $sparse[4] < members$. One must also check that $dense[sparse[4]] = 4$.

When inserting $v$ using the modified operations, as illustrated in Figure 8(c), we swap the values of $sparse[v]$ and $sparse[dense[members]]$, and likewise switch the values of $dense[members]$ and $dense[sparse[v]]$. This maintains the property that $v \in S \Leftrightarrow sparse[v] < members$.

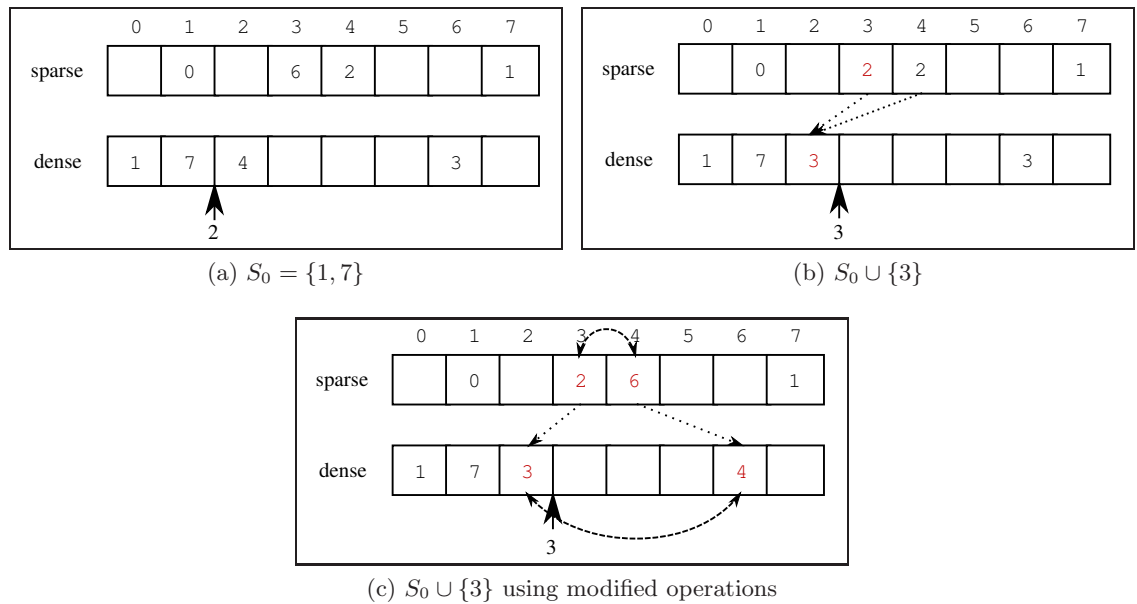

(a) $S_0 = \{1, 7\}$

(b) $S_0 \cup \{3\}$

(c) $S_0 \cup \{3\}$ using modified operations

Figure 8: A sparse representation for sets. (a) A possible state of the data structure representing $S_0 = \{1, 7\}$. (b) Inserting 3 into the data structure using the standard operations. *sparse*[3] is updated to point to the next element of *dense*, and the corresponding entry in *dense* points back to 3. Notably, both *sparse*[3] and *sparse*[4] now point to *dense*[2]. (c) Inserting 3 into the data structure using the modified operations. After the operation, both the *sparse* and *dense* arrays are maintained such that $\forall v \; dense[sparse[v]] = v$.

Dead subgraph memoization comes with a space cost of $O(|r|)$ to store the failure set *fset*. It reduces the time complexity of bddprop to $O((|r| - |fset|) \times |V|)$ and $O(|r| - |fset| + |V|)$ in practice.

#### 4.2.2 SHORTCUTTING

Shortcutting is an optimization to propagation on the BDD which notices that if all values in the current domains of variables $v_i, v_{i+1}, \cdots, v_N$ are fully supported, then we do not need to examine the rest of the nodes involving those variables. We keep a *high water mark hwater* which shows the least variable all of whose values are supported. If we ever reach a node numbered at or below the high water mark we only need to prove that it reaches $\mathcal{T}$, we do not need to fully explore the sub-graph below it.

A modified propagation algorithm taking into account shortcutting (and dead subgraph minimization) is given in Figures 9 and 10. The high water mark *hwater* is originally larger than the greatest variable appearing in the BDD.

The principle difference of imp_bddp is that if we reach a node with variable at or below the high water mark we use the simplified form shortcut_bddp which only checks whether the node can reach $\mathcal{T}$. The only other complexity is to update the high water mark *hwater* when we find all values of $v$ are supported ($E[v] = E'[v]$). shortcut_bddp has to be careful to mark all variables in nodes visited that reach $\mathcal{T}$ as mattering to the propagator.

```
imp_bddp(node,V,E) {
    if (in_set(fset, node)) return (1, 0);
    switch node {
    F: return (1,0);
    T: return (0,1);
    n(v, f, t):
        if (visit[node] ≥ time) return save[node];
        if (v ≥ hwater) return shortcut_bddp(node, V, E);
        reachf = 0; reacht = 0; maxvar = v;
        if (0 ∈ E[v]) {
            (rf0, rt0) = imp_bddp(f, V, E);
            reachf = rf0; reacht = rt0;
            if (rt0) {
                maxvar = top(f, V);
                E'[v] = E'[v] ∪ 0;
                if (hwater ≤ top(f, V) ∧ E'[v] == E[v]) {
                    hwater = v;
                    reachf = 1;
                    goto cleanup;
                }
            }
        }
        if (1 ∈ E[v]) {
            (rf1, rt1) = imp_bddp(t, V, E);
            reachf = reachf ∨ rf1; reacht = reacht ∨ rt1;
            if (rt1) {
                maxvar = max(maxvar, top(t, V));
                E'[v] = E'[v] ∪ 1;
                if (hwater ≤ top(t, V) ∧ E'[v] == E[v]) {
                    hwater = v;
                }
            }
        }
        if (¬reacht):
            insert(fset, node);
cleanup:
        for (v' ∈ V, v ≺ v' ≺ maxvar)
            E'[v] = E[v];
        if (reachf ∧ reacht) matters[v] = time;
        save[node] = (reachf, reacht);
        visit[node] = time;
        return (reachf, reacht);
    }
}
```

Figure 9: Pseudo-code for processing the constraint graph during propagation, using dead-subgraph memoization and shortcutting.

**Example 6** Consider the BDD for the constraint $|y \cap z| = 1$ when $N = 3$ shown in Figure 11(a). As no variables are fixed, we first explore the false paths, and find the $\mathcal{T}$ node. This provides complete support for $y_2, x_3, y_3$, so the high-water mark is updated to $y_2$. When searching for support for $x_2$ false, we no longer need to find support for anything beneath the high-water mark – we need only find a single path to true from the node labelled $y_2$. The high water mark then increases to $y_1$. Likewise, when finding support for $x_1$, everything below that point is already supported, so we explore only the first path to $\mathcal{T}$. The edges explored are shown doubled in 11(b).

Example 6 also illustrates that the impact of shortcutting is highly dependent on the order in which branches are searched, and the structure of the constraint – if we were to

```
shortcut_bddp(node,V,E) {
    if (in_set(fset, node)) return (1,0);
    switch node {
    T: return (0,1);
    n(v, f, t):
        rf0 = 0;
        if (visit[node] ≥ time) return save[node];
        if (0 ∈ E[v]) {
            (rf0, rt0) = shortcut_bddp(f, V, E);
            if (rt0) {
                if (1 ∈ E[v]) { matters[v] = time; rf0 = 1; }
                visit[node] = time; save[node] = (rf0, 1);
                return save[node];
            }
        }
        if (1 ∈ E[v]) {
            (rf1, rt1) = shortcut_bddp(t, V, E);
            if (rt1) {
                if (rf0) { matters[v] = time; rf1 = 1; }
                visit[node] = time; save[node] = (rf1, 1);
                return save[node];
            }
        }
        insert(fset, node);
        return (1, 0);
    }
}
```

Figure 10: Pseudo-code for the shortcut phase.

explore the *true* branches first, rather than the *false* branches, we would need to explore all nodes to find support for all variables. Clearly shortcutting does not change the asymptotic time or space complexity of the algorithm. Note that shortcutting for BDDs is more complex than the approach used by Cheng and Yap (2008) since they do not treat "long arcs" in MDDs.

## 5. Hybrid SAT Solver

Despite very fast propagation, a pure set bounds-based solver nevertheless suffers from an inability to analyze the reasons for failure, which results in repeated exploration of similar dead subtrees. This limits the performance of the solver on many hard problem instances.

In order to address this, we construct a hybrid solver which embeds BDD-based set bounds propagators within an efficient SAT solver. Search and conflict analysis are per-formed in the SAT solver, and the BDD propagators are used to generate inferences and clauses for the SAT solver to use during propagation.

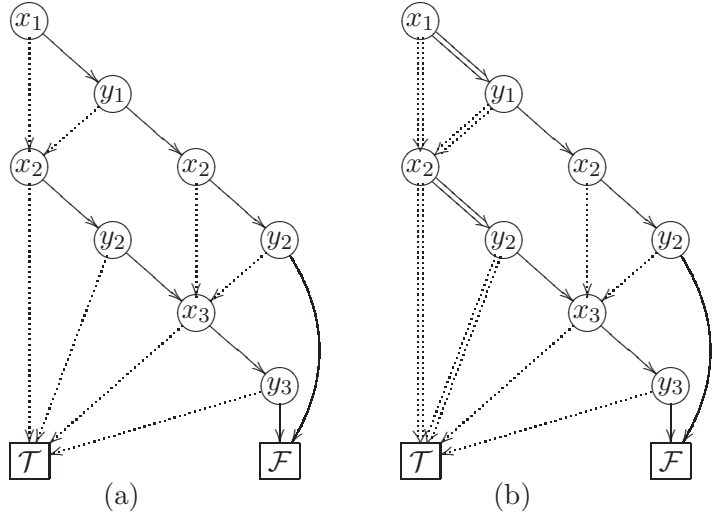

Figure 11: (a) The BDD representing $|x \cap y| \leq 1$ where $N = 3$. A node $n(v, f, t)$ is shown as a circle around $v$ with a dotted arrow to $f$ and full arrow to $t$. (b) The edges traversed by imp_bddp, when $E[v] = \{0, 1\}$ for all $v$, are shown doubled.

## 5.1 Efficient Reason Generation

Key to a successful SAT solver is the recording of *nogoods*, small subsets of the current variable assignments which independently result in failure. This allows similar subtrees to be eliminated from consideration, hence significantly reducing the search space.

In order to construct nogoods, it is necessary to explain the reason why each literal was set. in order to determine the chain of reasoning which resulted in a contradiction. In a pure SAT solver this is easy, as each variable is either a decision variable, or associated with a clause that caused propagation.

BDD-based propagation methods, however, do not automatically provide explanations for inference. The naive approach for generating a reason clause for a BDD inference is to enumerate all the fixed variables which occur in the propagator, and construct a clause from the negations:

$$\bigwedge_{l_i \in fix(\mathbf{B})} l_i \vDash l \Leftrightarrow l \vee \bigvee_{l_i \in fix(\mathbf{B})} \neg l_i$$

Unfortunately, this often results in very large reason clauses, particularly in the case of merged propagators or global constraints. As smaller clauses result in stronger nogoods being generated by the SAT solver, it is preferable to determine the minimal set of variables required to cause propagation, and include only those variables in the clause.

A method for constructing such minimal clauses was demonstrated by Hawkins and Stuckey (2006), but this method involves constructing new BDDs, eliminating redundant variables until the minimal BDD is constructed, then reading off the variables remaining in the BDD. Given the propagation algorithm herein avoids expensive BDD operations, we do not wish to use them for explanation.

Given that a set of assignments $\{l_0, \ldots, l_k\}$ entail a literal $l$ with respect to a constraint $C$, it is also true that

$$C \wedge \neg l \left( \bigwedge_{i \in \{0 \ldots k\}} l_i \right) \vDash \bot$$

As a result, the problem of finding a minimal reason for a given inference from a BDD is equivalent to fixing $\neg l$ and unfixing as many variables as possible without rendering $\mathcal{T}$ reachable.

The algorithm presented by Subbarayan (2008) provides a method to do this by traversing a static graph, again avoiding the need to construct intermediate BDDs. The algorithm, given in Figure 12, traverses each node $n(v, f, t)$ in a top-down memoing manner. At each node, it records if, given to the current domain, the $\mathcal{T}$ node is reachable. If the variable $v$ has been assigned a value, it also records if $\mathcal{T}$ is reachable from the conflicting edge; any such edges must not become relaxed, otherwise the partial assignment is no longer a conflict.

The graph is then traversed a second time, this time in a breadth-first manner. For each variable $v$, if all nodes which have been reached corresponding to $v$ variable may be relaxed without opening a path to $\mathcal{T}$, the $v$ is unfixed. If the $v$ remains fixed, $v$ is marked as part of the reason, and only the node corresponding to the value of $v$ is marked as reachable. Otherwise, $v$ is not in the minimal reason, and both the $f$ and $t$ nodes are marked as reached. The procedure returns the reason as a clause. The procedure is $O(|r|)$ in time and space complexity, but note this is $O(|r|)$ per new propagation that has to be explained!

**Example 7** Consider the constraint and assignments obtained in Example 4. It was determined that $E[y_1] = \{1\} \wedge E[z_2] = \{1\} \rightarrow E[x_2] = \{1\}$ (or equivalently, $1 \in y \wedge 2 \in z \rightarrow 2 \in x$). As such, the naive reason clause to explain $2 \in x$ would be $\neg y_1 \vee \neg z_2 \vee x_2$; however, it is possible to construct a smaller clause than this.

In order to construct the minimal reason for $E[x_2] = \{1\}$, we first set $E[x_2] = \{0\}$. The corresponding graph is shown in Figure 13(a), with nodes that are consistent with the partial assignment shown doubled. Note that as the solid edge from $x_2$ is not consistent with the assignment, $\mathcal{T}$ is not reachable along a doubled path from the root node.

The algorithm then determines the set of nodes from which $\mathcal{T}$ is reachable – these nodes shown doubled in Figure 13(b). These nodes must remain unreachable along the final reason; as such, the nodes which must remain fixed are the $x_2$ node and the leftmost $z_2$ node.

Finally, the algorithm progressively unfixes any variables which would not provide a path to $\mathcal{T}$ (in this case, $y_1$). The final path is shown in Figure 13(c), the resulting inference being $E[z_2] = \{1\} \rightarrow E[x_2] = \{1\}$; the corresponding reason clause is $x_2 \vee \neg z_2$.

## 5.2 Lazy Reason Generation

The simplest way to use reason generation is a so called *eager generation*, where whenever a BDD propagator makes a new inference, a minimal reason clause is generated and added to

```
construct_reason(r,V,D,var,sign) {
    Let r = n(v, t, f)
    D_old = D[var];
    D[var] = {1 − sign};
    forall (nodes n ∈ r) visit[n] := ⊥
    mark_reason(r,V,D);
    reached[v] = {r};
    if (sign)
        reason = var;
    else
        reason = ¬var;
    for (v′ ∈ V) {
        fixedvar = false;
        for (n ∈ reached[v′]) {
            fixedvar = fixedvar ∨ fixed[n];
        }
        if (fixedvar ∧ v′ ≠ var) {
            if (0 ∈ D[v])
                reason = reason ∨ v;
            else
                reason = reason ∨ ¬v;
        }
        for (n(v_n, f_n, t_n) ∈ reached[v′]) {
            if (¬fixedvar ∨ 1 ∈ D[v′])
                reached[v_n] = reached[v_n] ∪ t_n;
            if (¬fixedvar ∨ 0 ∈ D[v′])
                reached[v_n] = reached[v_n] ∪ f_n;
        }
    }
    D[var] = D_old;
    return reason;
}
```

```
mark_reason(node,V,D) {
    if (visit[node] ≠ ⊥) return visit[node];
    Let node = n(v, t, f)
    reachhi = mark_reason(t, V, D);
    reachlow = mark_reason(f, V, D);
    reacht = false;
    if (0 ∈ D[v])
        reacht = reacht ∨ reachlow;
    else
        fixed[node] = reachlow;
    if (1 ∈ D[v])
        reacht = reacht ∨ reachhi;
    else
        fixed[node] = reachhi;
    visit[node] = reacht;
    return reacht;
}
```

Figure 12: Pseudo-code for the reason generation algorithm by Subbarayan (2008). Constructs a minimal set of variables required to cause the inference $var = sign$.

the SAT solver. These clauses, however, cannot make any meaningful contribution to search until a conflict is detected – they cannot cause any propagation until the solver backtracks beyond the fixed variable, and no conflict clauses are constructed until there is a conflict. As there is a degree of overhead in adding and maintaining a large set of these clauses in the solver, it may be better to delay constructing these reasons until they are actually required to explain a conflict.

We can instead apply the reason generation only when the SAT conflict analysis asks for the explanation of a literal set by the BDD solver. We call this *lazy generation*. In order to do so, we must determine the state of the propagator which caused the inference. We implement this by recording the order in which literals become fixed in a propagator. When generating a reason for a variable $v$ becoming fixed, we look at each variable in the propagator, and unfix any variable $v'$ such that $time(v) \leq time(v')$, then restore them after the reason is constructed.

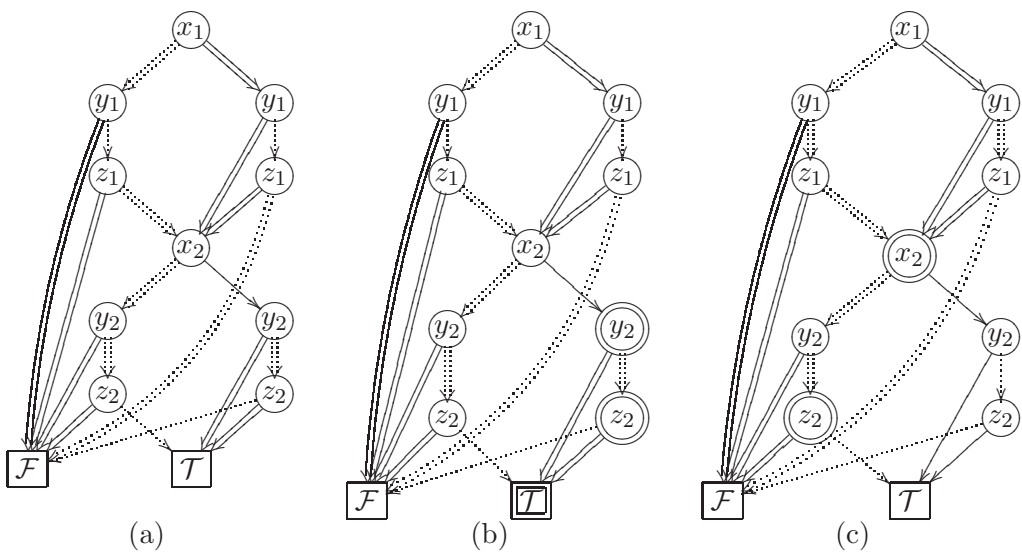

Figure 13: (a) The BDD representing $x = y \cup z$ where $N = 2$, with $E[y_1] = \{1\}$, $E[z_2] = \{1\}$ and $E[x_2] = \{0\}$. Edges consistent with the partial assignment are shown doubled. (b) Nodes which must remain unreachable in the reason are shown doubled. (c) Edges reachable along the minimal reason are shown doubled, as are nodes which remain fixed.

### 5.3 Hybrid Architecture

The hybrid SAT solver embeds BDD propagators inside the SAT engine. The architecture is illustrated in Figure 14. The usual SAT engine architecture is shown on the left. BDD propagation is added as shown on the right. Unit propagation causes Boolean literals to be fixed which may require that BDD propagators need to be awoken. We attach to each Boolean variable representing part of a set variable $x$ the BDD propagators involving that set variable. When unit propagation reaches a fixpoint, the trail of fixed literals is traversed and each BDD propagator that includes one of these literals is scheduled for execution. If we are using filtering, it is only scheduled if the literal is one which matters to the propagator. Then we execute the scheduled BDD propagators using `imp_bddp`. If the BDD propagator fixes some literals then these are added to the trail of the unit propagation engine. If we are using eager reason generation then we also immediately build a clause explaining the propagation and add it to the clause database and record this clause as the reason for the propagation of the literal.

If we are using lazy reason generation, instead we record as the reason simply a pointer to the BDD propagator which causes the literal to be fixed. Then if conflict analysis demands an explanation for the literal, we call the reason generation for the BDD propagator, using the state at the time when the literal was fixed, to build an explaining clause. This is used in conflict analysis. We replace the reason for the literal in the trail by the generated explanation clause and also add the explanation clause to the database.

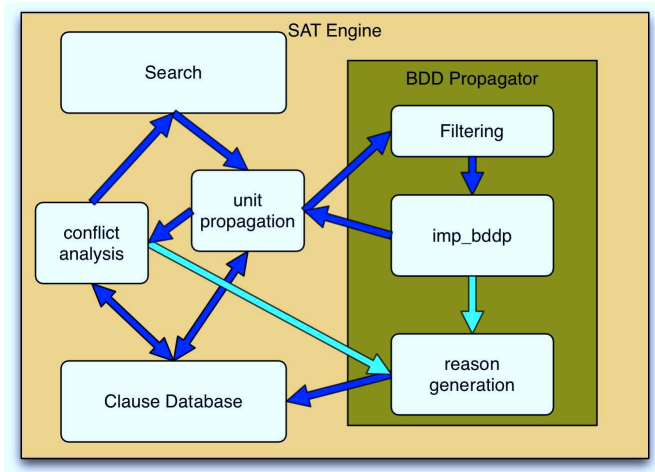

Figure 14: Architecture for the hybrid BDD-SAT solver.

The implementation inherits almost all features of the underlying SAT solver. Eager reason clauses are added as nogoods, and deleted when the SAT solver decides to eliminate nogoods, lazy reason clauses are only generated on demand during conflict analysis. They are added to the clause database even though this is not necessary, since its makes memoing which explanations have been already performed simpler. The hybrid solver can make use of restarting activity based search, and restarts, although we also extend the search capabilities to allow some simple static searches as these can be preferable for the set problems we tackle.

## 6. Experimental Results

We built a hybrid SAT solver implementing the algorithms described above. The solver is based on MiniSAT 2.0 (dated 070721) (Eén & Sörensson, 2003), which has been modified to include the BDD-based propagation engine. BDDs are constructed using the BuDDy BDD package (http://sourceforge.net/projects/buddy/) All BDDs are constructed at the beginning of execution, then converted to the static graph used during propagation. Indeed, for many of the smaller problems solved in Section 6, the majority of the solution time is used in constructing the BDDs.

The BDD propagators are executed at a lower priority level than unit propagation, in order to detect conflict as early as possible. Reason clauses which are generated by the set-bounds propagator are added to the SAT solver as learnt clauses, as otherwise the number of clauses added to the solver during propagation of hard problems can overwhelm the solver.

Experiments were conducted on a 3.00GHz Core2 Duo with 2 Gb of RAM running Ubuntu GNU/Linux 8.10. All problems were terminated if not completed within 10 minutes.

We experimented on 3 classes of set benchmarks: social golfers, Steiner systems, and Hamming codes. Unless otherwise specified, the hybrid solver is always executed using lazy reason generation.

We compare with the Gecode 3.1.0 set bounds propagation solver since it is acknowledged as one of the fastest solvers available, as well as ECLiPSe 6.0 #100. We also compare with published results of the CARDINAL (Azevedo, 2007) and LENGTH-LEX (Yip & Van Hentenryck, 2009) solvers on the same problems.

## 6.1 Social Golfers

A common set benchmark is the "Social Golfers" problem, which consists of arranging $N = g \times s$ golfers into $g$ groups of $s$ players for each of $w$ weeks, such that no two players play together more than once. Again, we use the same model as used by Lagoon and Stuckey (2004), using a $w \times g$ matrix of set variables $v_{ij}$ where $1 \leq i \leq w$ and $1 \leq j \leq g$.

$$\left(\bigwedge_{i=1}^{w} \texttt{partition}^{<}(v_{i1}, \ldots, v_{ig})\right) \ \wedge \ \left(\bigwedge_{i=1}^{w} \bigwedge_{j=1}^{g} |v_{ij}| = s\right) \ \wedge$$
$$\left(\bigwedge_{i,j \in \{1...w\}, \ i \neq j} \ \bigwedge_{k,l \in \{1...g\}} |v_{ik} \cap v_{jl}| \leq 1\right) \wedge \left(\bigwedge_{i=1}^{w-1} \bigwedge_{j=i+1}^{w} v_{i1} \leq v_{j1}\right)$$

The global constraint $\texttt{partition}^{<}$ ensures its arguments are pairwise disjoint and imposes a lexicographic order on its arguments, i.e. $v_{i1} < \cdots < v_{ig}$. The corresponding propagator is based on a single BDD. We construct BDD propagators for each of the constraint forms $|v \cap v'| \leq 1$, $v \leq v'$ and $|v| = s$. Note that the first form would typically be decomposed into $u = v \cap v' \wedge |u| \leq 1$ in a normal set bounds propagator.

The hybrid solver constructs one BDD for each of the 4 terms in the above equation, instantiating constraints accordingly.

Table 1 shows the results using a static search strategy on easy problems. The search fixes the elements of the sets $v_{ij}$ in order $v_{11}, v_{12}, \ldots, v_{1g}, v_{21}, \ldots, v_{wg}$, always trying to first place the least element in the set then excluding it from the set. We compare against the reported results for the original BDD-SAT hybrid solver of Hawkins and Stuckey (2006) versus a number of variations of our hybrid. `base` is the base solver of Figures 4 and 5, while +f indicates with filtering of Section 4.1 added, +s indicates with dead subgraph memoization and shortcutting added (Section 4.2) using the original sparse set code, +i is these optimizations with the improved sparse set code. We also combine filtering with the other optimizations. The table shows time and number of fails for each variant, where the solvers with identical failure behaviour are grouped together. Note that filtering can change the search by reordering the propagations and hence changing the nogoods that are generated, while the other optimizations cannot except that shortcutting can change the results of filtering (and hence change search). While filtering improves on the base line, dead subgraph memoization and shortcutting do not, although we can see the benefit of the improved sparse set operations. Comparing against the solver of Hawkins and Stuckey (2006), which was run on a (♯) 2.4GHz Pentium 4, we find that, slightly different number of backtracks and slightly faster machine not withstanding, the solver presented here is roughly an order of magnitude faster.

Table 2 shows the results using VSIDS search on easy problems. It compares against the solver of Hawkins and Stuckey (2006) and a Tseitin decomposition. The results are the same as for Table 1, and overall VSIDS is better than static search. The table illustrates some of the difficulty of comparing systems using VSIDS search, since small differences can drastically change the search space. The solver +f is the best except for a bad-performance on 7,5,3. The base solver is around 5 times faster per failure than the solver of Hawkins and Stuckey (2006). The Tseitin decomposition is not competitive, even if we discount the results on 7,5,3.

| Problem | Static Search | | | | | | | | | | | |
|---|---|---|---|---|---|---|---|---|---|---|---|---|
| | Hawkins[2] | | Hybrid | | | | | | | | | |
| | | | base | +s | +i | fails | +f | fails | +fs | +fi | fails |
| | time | fails | | | | | | | | | | |
| 2,5,4 | 0.10 | 11 | 0.03 | **0.02** | **0.02** | 19 | **0.02** | 19 | **0.02** | **0.02** | 19 |
| 2,6,4 | 0.10 | 45 | 0.04 | 0.04 | 0.05 | 126 | 0.05 | 153 | 0.05 | 0.04 | 153 |
| 2,7,4 | 0.20 | 90 | 0.06 | 0.07 | 0.07 | 148 | 0.07 | 265 | 0.07 | 0.07 | 265 |
| 2,8,5 | 0.80 | 472 | 2.84 | 3.15 | 3.13 | 8856 | 0.47 | 1119 | 0.50 | 0.50 | 1119 |
| 3,5,4 | 0.10 | 11 | **0.02** | **0.02** | 0.04 | 19 | **0.02** | 19 | 0.03 | **0.02** | 19 |
| 3,6,4 | 0.20 | 48 | **0.04** | 0.05 | 0.07 | 129 | 0.05 | 156 | 0.06 | 0.06 | 156 |
| 3,7,4 | 0.70 | 81 | 0.12 | 0.08 | 0.11 | 165 | 0.10 | 282 | 0.14 | 0.12 | 282 |
| 4,5,4 | 0.20 | 11 | 0.03 | 0.03 | 0.04 | 19 | **0.02** | 19 | **0.02** | 0.03 | 19 |
| 4,6,5 | 0.70 | 81 | 0.25 | 0.27 | 0.26 | 559 | **0.07** | 77 | 0.10 | 0.09 | 77 |
| 4,7,4 | 0.80 | 105 | 0.11 | 0.14 | 0.15 | 171 | 0.18 | 288 | 0.17 | 0.17 | 288 |
| 4,9,4 | 1.90 | 32 | 0.18 | 0.18 | 0.18 | 40 | 0.14 | 40 | **0.14** | **0.14** | 40 |
| 5,4,3 ⋆ | 12.00 | 9568 | 2.58 | 3.00 | 2.92 | 10294 | 2.35 | 10209 | 2.69 | 2.69 | 10188 |
| 5,5,4 | 2.30 | 1167 | 0.42 | 0.48 | 0.46 | 1328 | **0.33** | 1293 | 0.40 | 0.36 | 1297 |
| 5,7,4 | 1.50 | 159 | 0.18 | 0.25 | 0.21 | 217 | 0.24 | 335 | 0.25 | 0.24 | 335 |
| 5,8,3 | 0.90 | 12 | **0.06** | 0.10 | 0.07 | 10 | 0.08 | 10 | **0.06** | 0.10 | 10 |
| 6,4,3 ⋆ | 2.10 | 908 | 0.51 | 0.60 | 0.57 | 1699 | 0.33 | 1079 | 0.32 | 0.33 | 922 |
| 6,5,3 | 0.90 | 282 | 0.13 | 0.14 | 0.16 | 278 | **0.09** | 261 | 0.11 | 0.14 | 257 |
| 6,6,3 | 0.40 | 5 | 0.05 | 0.04 | 0.05 | 5 | **0.03** | 5 | **0.03** | 0.04 | 5 |
| 7,5,3 | 18.20 | 6152 | 3.79 | 4.67 | 4.45 | 7616 | **2.10** | 5702 | 3.08 | 2.97 | 6302 |
| 7,5,5 ⋆ | 0.80 | 100 | 0.20 | 0.20 | 0.18 | 121 | 0.18 | 121 | 0.21 | 0.20 | 121 |
| Total | 44.90 | 19340 | 11.64 | 13.53 | 13.19 | 31819 | 6.92 | 21452 | 8.45 | 8.33 | 21874 |

Table 1: First-solution performance results on the Social Golfers problem using a static, first-element in set ordering. Instances marked with (⋆) are unsatisfiable, entries marked with '—' did not complete within 10 minutes.

For social golfers, dead-subset memoization and shortcutting provide no advantage (when we discount the drastically different search for 7,5,3 using VSIDS). While the number of nodes processed can be reduced slightly, this is not enough to repay the additional cost of computation at each node.

Table 3 compares the reason generation strategies: *eager* reasoning which constructs reasons as soon as inference is detected; and *lazy* reasoning which only those reasons necessary to determine the first UIP or perform conflict clause minimization.

Table 3 compares the base solver with and without filtering (since dead subgraph memoization and shortcutting do not help here) on harder social golfer problems using a static search. It shows time (base) as well as the number of reasons generated and fails in order to find a first solution. For these harder examples filtering is highly beneficial. Here we can see that the number of reasons generated by lazy reasoning is about half of that required by eager reasoning, but it doesn't make that much difference to the computation time, since propagation dominates the time spent in the solver. Interestingly not adding reasons eagerly also seems to generate slightly better nogoods as the search is usually smaller. Table 4 shows the results using VSIDS search on these harder instances. It appears the advantages of lazy reasoning are increased by the use of VSIDS, presumably because the better nogoods are then more useful in driving search.

| Problem | VSIDS Search | | | | | | | | | | | | |
| | Hawkins[2] | | Hybrid | | | | | | | | | Tseitin | |
| | time | fails | base | +s | +i | fails | +f | fails | +fs | +fi | fails | time | fails |
| 2,5,4 | 0.10 | 22 | 0.04 | 0.03 | 0.03 | 4 | **0.02** | 4 | **0.02** | 0.03 | 4 | 0.03 | 7 |
| 2,6,4 | 0.10 | 64 | **0.02** | 0.03 | **0.02** | 20 | 0.03 | 20 | **0.02** | **0.02** | 20 | 0.04 | 37 |
| 2,7,4 | 0.20 | 119 | **0.03** | **0.03** | **0.03** | 13 | 0.04 | 13 | 0.05 | 0.04 | 13 | 0.06 | 55 |
| 2,8,5 | 1.30 | 622 | 0.10 | 0.12 | 0.11 | 109 | 0.10 | 109 | 0.10 | 0.10 | 109 | **0.09** | 78 |
| 3,5,4 | 0.10 | 24 | 0.04 | 0.04 | **0.02** | 51 | 0.03 | 51 | 0.03 | 0.04 | 51 | 0.05 | 170 |
| 3,6,4 | 0.30 | 58 | 0.05 | **0.04** | **0.04** | 80 | 0.06 | 80 | **0.04** | **0.04** | 80 | 0.07 | 268 |
| 3,7,4 | 0.60 | 92 | **0.06** | **0.06** | **0.06** | 78 | 0.07 | 79 | **0.06** | 0.10 | 79 | 0.12 | 469 |
| 4,5,4 | 0.40 | 122 | 0.05 | 0.06 | 0.06 | 108 | 0.04 | 116 | 0.06 | 0.06 | 116 | 0.14 | 1143 |
| 4,6,5 | 1.30 | 304 | 0.26 | 0.26 | 0.26 | 309 | 0.13 | 158 | 0.14 | 0.15 | 205 | 0.49 | 3156 |
| 4,7,4 | 1.00 | 98 | 0.09 | 0.11 | 0.10 | 102 | 0.10 | 103 | 0.09 | **0.08** | 103 | 0.25 | 1020 |
| 4,9,4 | 2.00 | 59 | 0.16 | 0.18 | 0.18 | 36 | **0.14** | 36 | 0.18 | 0.15 | 36 | 0.63 | 1037 |
| 5,4,3 ★ | 5.60 | 5876 | 1.23 | 1.42 | 1.35 | 5869 | **0.56** | 3139 | 0.69 | 0.67 | 3184 | 4.74 | 26769 |
| 5,5,4 | 1.90 | 581 | 4.14 | 5.16 | 4.80 | 9846 | 0.91 | 2487 | 0.77 | 0.74 | 1754 | 0.58 | 3475 |
| 5,7,4 | 1.50 | 104 | 0.16 | 0.13 | 0.13 | 77 | **0.11** | 84 | 0.13 | 0.12 | 84 | 1.16 | 3596 |
| 5,8,3 | 1.70 | 425 | 0.08 | 0.10 | 0.10 | 29 | 0.10 | 29 | 0.10 | 0.10 | 29 | 0.52 | 918 |
| 6,4,3 ★ | 0.20 | 71 | 0.18 | 0.17 | 0.17 | 425 | **0.14** | 479 | 0.30 | 0.28 | 1013 | 2.83 | 17595 |
| 6,5,3 | 4.30 | 2801 | 0.25 | 0.27 | 0.29 | 369 | 0.18 | 409 | 0.17 | 0.16 | 397 | 1.85 | 8675 |
| 6,6,3 | 1.00 | 275 | 0.07 | 0.06 | 0.07 | 36 | 0.07 | 70 | 0.08 | 0.09 | 70 | 1.09 | 3547 |
| 7,5,3 | 18.00 | 7018 | 8.81 | 11.08 | 10.72 | 18949 | 39.35 | 93789 | 2.47 | 2.38 | 4554 | 45.54 | 77786 |
| 7,5,5 ★ | 2.00 | 139 | 0.14 | 0.11 | 0.12 | 47 | **0.10** | 47 | 0.13 | **0.10** | 47 | 0.93 | 1977 |
| Total | 43.60 | 18874 | 15.96 | 19.46 | 18.66 | 36557 | 42.28 | 101302 | 5.63 | **5.45** | 11948 | 61.21 | 151778 |

Table 2: First-solution performance results on the Social Golfers problem using a VSIDS search strategy.

| Problem | Social Golfers | | | | | | | | | | | |
| | Lazy Reason Generation | | | | | | Eager Reason Generation | | | | | |
| | base | reasons | fails | +f | reasons | fails | base | reasons | fails | +f | reasons | fails |
| 7,5,3 | 6.34 | 62630 | 13071 | **5.49** | 65447 | 13079 | 8.76 | 117323 | 13273 | 7.38 | 117657 | 12598 |
| 2,6,5 | 0.14 | 1673 | 581 | **0.03** | 317 | 66 | 0.17 | 3026 | 581 | 0.04 | 740 | 66 |
| 4,6,5 | 0.46 | 7058 | 1067 | **0.38** | 7026 | 1037 | 0.55 | 11833 | 1066 | 0.48 | 11967 | 1028 |
| 6,10,3 | 0.90 | 6675 | 871 | **0.72** | 6973 | 946 | 1.04 | 9942 | 820 | 0.87 | 10101 | 849 |
| 9,10,3 | 16.91 | 15522 | 3857 | **15.74** | 15103 | 3708 | 27.88 | 34181 | 4039 | 25.77 | 32945 | 3853 |
| 10,10,3 | **109.43** | 28110 | 11462 | 130.04 | 32580 | 13128 | 198.90 | 81270 | 12755 | 187.11 | 79461 | 12338 |
| Total | **134.18** | 121668 | 30909 | 152.40 | 127446 | 31964 | 237.30 | 257575 | 32534 | 221.65 | 252871 | 30732 |

Table 3: First-solution performance results on harder Social Golfers problems, using a static least-element in set search method. Results are given comparing eager and lazy reason generation.

Finally Table 5 compares against a number of different systems. We use the model of social-golfers described in the work of Yip and Van Hentenryck (2009), which in addition fixes the first week, and the first group of the second week to eliminate symmetric solutions. We use the instances reported by Yip and Van Hentenryck (2009). We show results for our base solver with and without filtering. We compare against Gecode 3.1.0 and Eclipse 6.0 #100, both which implement a set bounds propagation combined with limited cardinality reasoning, on an identical MiniZinc model of social-golfers running on our 3GHz Core2Duo. Gecode arguably represents the state of the art for set bounds propagation solving. We also compare against the published results of the Cardinal solver (Azevedo, 2007), which uses more complex cardinality reasoning for set solving, using a (†) Pentium 4 2.4GHz machine, and the recently published results for the Length-Lex solver of Yip and Van Hentenryck (2009), which maintains bounds on sets variables in terms of the length-lex order (see Gervet & Van Hentenryck, 2006; Yip & Van Hentenryck, 2009 for details) running on a (‡) C2D-M 2.53GHz machine. The pure set bounds solvers cannot compete with our approach since the search space without using nogood recording is just too big. None of the other systems except Length-lex can solve all of these instances. One can a see a drastic difference between number of failures for Gecode, which uses set bounds propagation without learning, versus our base solver. Gecode can sometimes require less failures on easy problems since it combines cardinality reasoning with bounds reasoning, but on hard problems the advantages of learning to prune similar searches in other parts of the tree dominates completely. The stronger pruning of Length-lex compared to set bounds means it can often improve on fails compared to base but learning is more robust. The hybrid solver is overall around an order of magnitude faster than Length-Lex.

| Problem | Social Golfers | | | | | | | | | | | |
|---------|------|---------|-------|------|---------|-------|------|---------|-------|------|---------|-------|
| | Lazy Reason Generation | | | | | | Eager Reason Generation | | | | | |
| | base | reasons | fails | +f | reasons | fails | base | reasons | fails | +f | reasons | fails |
| 7,5,3 | **0.20** | 2096 | 217 | 1.03 | 12638 | 2608 | 0.22 | 4328 | 212 | 2.18 | 63780 | 4911 |
| 2,6,5 | **0.02** | 38 | 4 | **0.02** | 38 | 4 | 0.03 | 350 | 4 | 0.04 | 350 | 4 |
| 4,6,5 | **0.06** | 176 | 18 | **0.06** | 176 | 18 | 1.28 | 28026 | 1565 | 0.08 | 1604 | 40 |
| 6,10,3 | 0.22 | 188 | 7 | **0.18** | 188 | 7 | 0.34 | 1824 | 7 | 0.34 | 1823 | 7 |
| 9,10,3 | 0.92 | 1743 | 110 | **0.45** | 666 | 68 | 1.59 | 6685 | 134 | 1.20 | 6310 | 107 |
| 10,10,3 | 1.51 | 1917 | 139 | **0.64** | 641 | 46 | 2.06 | 7707 | 200 | 1.08 | 4310 | 57 |
| Total | 2.93 | 6158 | 495 | **2.38** | 14347 | 2751 | 5.52 | 48920 | 2122 | 4.92 | 78177 | 5126 |

Table 4: First-solution performance results on harder Social Golfers problems, using a VSIDS search method. Results are given comparing eager and lazy reason generation.

## 6.2 Steiner Systems

Another commonly used benchmark for set constraint solvers is the calculation of small Steiner systems. A Steiner system $S(t, k, N)$ is a set $X$ of cardinality $N$ and a collection $C$ of subsets of $X$ of cardinality $k$ (called 'blocks'), such that any $t$ elements of $X$ are in exactly one block. Any Steiner system must have exactly $m = \binom{N}{t}/\binom{k}{t}$ blocks (Theorem 19.2 of van Lint & Wilson, 2001).

We model the Steiner problem similarly to Lagoon and Stuckey (2004) extended for the case of more general Steiner Systems. We model each block as a set variable $s_1, \ldots, s_m$, with the constraints:

$$\bigwedge_{i=1}^{m} (|s_i| = k) \ \wedge$$

$$\bigwedge_{i=1}^{m-1} \bigwedge_{j=i+1}^{m} (|s_i \cap s_j| \leq t - 1 \wedge s_i < s_j)$$

For comparison with the results of Azevedo (2007) and Yip and Van Hentenryck (2009), we construct a dual model with additional variables $d_1, \ldots, d_N$, with additional constraints

| Problem | GECODE | | ECLIPSE | CARDINAL† | LENGTH-LEX‡ | | base | | +f | |
|---|---|---|---|---|---|---|---|---|---|---|
| 4,4,2 | **0.00** | 0 | 0.56 | | 0.01 | 0 | 0.01 | 3 | 0.01 | 3 |
| 5,4,2 | **0.00** | 0 | 0.54 | | 0.01 | 0 | 0.02 | 4 | 0.02 | 4 |
| 6,4,2 | **0.00** | 0 | 0.55 | | 0.01 | 0 | 0.02 | 8 | 0.02 | 8 |
| 7,4,2 | **0.00** | 0 | 0.62 | | 0.01 | 0 | 0.02 | 13 | 0.04 | 13 |
| 4,4,3 | **0.00** | 6 | 0.56 | | 0.01 | 0 | 0.01 | 9 | 0.02 | 9 |
| 5,4,3 | 1.72 | 5781 | 11.21 | 165.63 | 0.40 | 732 | 0.46 | 1877 | **0.35** | 1799 |
| 6,4,3 | **0.01** | 45 | 0.66 | 94.67 | 0.02 | 29 | 0.03 | 57 | 0.04 | 57 |
| 4,4,4 | **0.00** | 0 | 0.57 | | 0.06 | 111 | 0.02 | 3 | 0.02 | 3 |
| 5,4,4 | **0.00** | 0 | 0.61 | | 0.05 | 57 | 0.02 | 8 | 0.03 | 8 |
| 3,5,2 | **0.00** | 1 | 0.54 | | 0.01 | 0 | 0.02 | 1 | 0.02 | 1 |
| 4,5,2 | **0.00** | 2 | 0.57 | | 0.01 | 0 | 0.01 | 3 | 0.03 | 3 |
| 5,5,2 | **0.00** | 2 | 0.61 | | 0.01 | 0 | 0.02 | 5 | 0.02 | 5 |
| 6,5,2 | **0.00** | 6 | 0.66 | | 0.01 | 0 | 0.01 | 8 | 0.03 | 8 |
| 7,5,2 | **0.01** | 17 | 0.71 | | 0.02 | 1 | 0.04 | 14 | 0.04 | 14 |
| 8,5,2 | **0.01** | 22 | 0.81 | | 0.02 | 1 | 0.06 | 24 | 0.05 | 24 |
| 9,5,2 | **0.01** | 17 | 0.86 | | 0.02 | 1 | 0.12 | 40 | 0.14 | 40 |
| 3,5,3 | **0.01** | 49 | 0.58 | | **0.01** | 1 | 0.02 | 10 | 0.02 | 10 |
| 4,5,3 | 0.03 | 73 | 0.65 | | **0.01** | 1 | 0.03 | 18 | **0.01** | 16 |
| 5,5,3 | 0.03 | 105 | 0.76 | | **0.02** | 5 | 0.03 | 30 | 0.03 | 30 |
| 6,5,3 | 110.33 | 335531 | — | — | **0.41** | 316 | 0.80 | 2516 | 0.56 | 2345 |
| 7,5,3 | — | — | — | — | 6.32 | 46117 | 6.32 | 13071 | **5.50** | 13079 |
| 2,5,4 | 0.09 | 1090 | 1.26 | | **0.01** | 11 | 0.02 | 21 | 0.02 | 21 |
| 3,5,4 | 0.11 | 605 | 1.10 | 1.89 | **0.02** | 24 | 0.03 | 36 | **0.02** | 36 |
| 4,5,4 | 0.09 | 298 | 0.98 | 3.13 | 0.14 | 194 | **0.07** | 191 | 0.08 | 189 |
| 5,5,4 | **0.19** | 410 | 1.40 | 28.65 | 1.87 | 1947 | 0.98 | 2592 | 0.74 | 2356 |
| 3,5,5 | **0.00** | 1 | 0.65 | | 0.06 | 93 | 0.02 | 0 | 0.02 | 0 |
| 4,5,5 | **0.01** | 13 | 0.74 | | 4.72 | 6876 | 0.05 | 29 | 0.04 | 27 |
| 5,5,5 | **0.04** | 45 | 0.84 | | 54.27 | 50623 | 0.07 | 64 | 0.06 | 64 |
| 6,5,5 | **0.03** | 30 | 0.96 | | 29.21 | 15769 | 0.15 | 167 | 0.15 | 221 |
| 7,5,5 | 8.11 | 12274 | 161.12 | — | **0.01** | 1 | 3.28 | 4728 | 8.17 | 11736 |
| 2,6,3 | **0.00** | 0 | 0.56 | | **0.00** | 0 | 0.02 | 4 | 0.02 | 4 |
| 3,6,3 | **0.01** | 30 | 0.69 | | **0.01** | 1 | 0.03 | 4 | **0.01** | 4 |
| 4,6,3 | **0.01** | 23 | 0.71 | | **0.01** | 1 | 0.04 | 11 | 0.02 | 11 |
| 5,6,3 | 0.09 | 311 | 1.08 | | **0.02** | 6 | 0.04 | 15 | 0.04 | 15 |
| 6,6,3 | 0.15 | 388 | 1.32 | 1.20 | **0.04** | 10 | 0.39 | 932 | 0.27 | 959 |
| 2,6,4 | 1.91 | 16608 | 8.12 | 1.75 | **0.01** | 14 | 0.04 | 58 | 0.02 | 58 |
| 3,6,4 | 3.68 | 15948 | 10.16 | 4.62 | **0.03** | 42 | 0.06 | 97 | 0.05 | 97 |
| 2,6,5 | — | — | — | | **0.05** | 118 | 0.13 | 581 | **0.05** | 66 |
| 3,6,5 | 547.91 | 1893577 | — | | 2.54 | 3351 | 0.15 | 532 | **0.08** | 178 |
| 4,6,5 | 275.38 | 584532 | — | — | 32.60 | 31270 | 0.45 | 1067 | **0.35** | 1037 |
| 5,6,5 | 96.89 | 145371 | 265.40 | | 28.76 | 6758 | 15.24 | 26495 | **2.85** | 7295 |
| 3,6,6 | **0.01** | 8 | 0.82 | | 0.82 | 661 | 0.04 | 0 | 0.03 | 0 |
| 2,7,2 | **0.00** | 1 | 0.60 | | 0.01 | 0 | 0.03 | 0 | 0.03 | 0 |
| 2,7,3 | **0.00** | 10 | 0.64 | | 0.01 | 1 | 0.02 | 0 | 0.03 | 0 |
| 2,7,4 | 5.64 | 39833 | 17.43 | 2.82 | **0.01** | 0 | 0.02 | 19 | 0.05 | 19 |
| 3,7,4 | 13.40 | 46621 | 27.23 | 6.37 | **0.03** | 21 | 0.07 | 69 | 0.07 | 72 |
| 4,7,4 | 10.54 | 24216 | 19.38 | 12.46 | **0.05** | 26 | 0.08 | 62 | 0.08 | 63 |
| 5,7,4 | 11.32 | 18785 | 17.17 | 17.18 | 0.36 | 152 | 0.45 | 243 | **0.16** | 234 |
| 2,7,5 | — | — | — | | 0.31 | 574 | 0.45 | 1944 | **0.07** | 133 |
| 2,7,6 | — | — | — | | 0.78 | 1271 | 1.44 | 6031 | **0.18** | 566 |
| 2,7,7 | **0.01** | 0 | 0.86 | | 0.28 | 0 | 0.06 | 0 | 0.05 | 0 |
| 5,8,3 | — | — | — | 1.01 | 34.52 | 45477 | **0.09** | 71 | 0.11 | 70 |
| 4,8,4 | 26.92 | 56844 | 46.03 | | **0.06** | 18 | 0.12 | 67 | 0.10 | 67 |
| 2,8,5 | — | — | — | — | 0.25 | 307 | 0.56 | 2145 | **0.23** | 291 |
| 4,9,4 | 8.93 | 11854 | 12.18 | 42.45 | 0.21 | 94 | **0.19** | 90 | **0.19** | 100 |
| 6,10,3 | — | — | — | | 5.86 | 2941 | 0.91 | 871 | **0.73** | 946 |
| 9,10,3 | — | — | — | | 233.80 | 45437 | 16.66 | 3857 | **15.64** | 3708 |
| 10,10,3 | — | — | — | | 210.80 | 25246 | **110.80** | 11462 | 129.66 | 13128 |
| 4,10,4 | 17.33 | 16345 | 17.10 | | **0.27** | 104 | 0.52 | 409 | 0.46 | 419 |
| 5,10,4 | 34.62 | 36294 | 46.58 | | **0.58** | 149 | 0.76 | 576 | 0.65 | 596 |
| Total | — | — | — | | 719.11 | 286960 | **162.39** | 83262 | 168.58 | 62265 |

Table 5: Comparison against solvers using different propagation mechanisms, using the model and instances described by Yip and Van Hentenryck (2009). "—" denotes failure to complete a test-case in 10 minutes (or 15 minutes for CARDINAL). A blank entry means there is no published result to compare.

as shown:

$$\bigwedge_{i=1}^{m} \bigwedge_{j=1}^{N} (j \in s_i \Leftrightarrow i \in d_j) \ \wedge$$

$$\bigwedge_{j=1}^{N} (|d_j| = \frac{m \times k}{N})$$

| Problem | Gecode | | Eclipse | Card[†] | Length-lex[‡] | | Static Hybrid | | VSIDS Hybrid | | Tseitin | |
|---|---|---|---|---|---|---|---|---|---|---|---|---|
| | time | fails | time | time | time | fails | base | fails | base | fails | time | fails |
| 2,3,7 | **0.00** | 0 | 0.53 | 0.01 | **0.00** | 0 | 0.01 | 0 | 0.03 | 5 | 0.04 | 59 |
| 2,3,9 | **0.00** | 3 | 0.56 | 0.05 | 0.01 | 1 | 0.02 | 1 | 0.02 | 17 | 0.81 | 3804 |
| 2,3,13 | 0.03 | 18 | 1.20 | 0.61 | 0.05 | 10 | 0.06 | 9 | **0.02** | 24 | 1.93 | 7879 |
| 2,3,15 | **0.04** | 0 | 2.15 | 0.91 | 0.09 | 0 | 0.07 | 0 | 0.32 | 295 | 143.81 | 124205 |
| 2,3,19 | 0.65 | 144 | 8.09 | 7.94 | 0.46 | 164 | 0.37 | 78 | **0.07** | 106 | 14.01 | 34089 |
| 2,3,21 | 2.61 | 413 | — | 39.07 | 1.04 | 448 | **0.82** | 225 | 39.19 | 42688 | — | — |
| 2,3,25 | — | — | — | — | 14.07 | 5100 | **7.10** | 2474 | — | — | — | — |
| 2,3,27 | — | — | — | — | 23.55 | 7066 | **12.88** | 3401 | 229.59 | 113373 | — | — |
| 2,3,31 | — | — | — | 48.52 | **5.29** | 0 | 5.38 | 0 | — | — | — | — |
| 2,3,33 | — | — | — | — | 443.07 | 111923 | | | **19.30** | 8228 | — | — |

Table 6: First-solution performance results on the alternate Steiner Systems instances using a dual model. Gecode and the sequential hybrid use a sequential least-element in set search strategy over the dual variables. The VSIDS hybrid and Tseitin decomposition use VSIDS search. "—" denotes failure to complete a test-case in 10 minutes due to either timeout or a memory error. A blank entry means there is no published result to compare.

We create BDD propagators for each of the the constraint forms $|v| = \frac{m \times k}{N}$ and $|v \cap v'| \le t - 1 \wedge v < v' \wedge |v| = k \wedge |v'| = k$. Again note in non-BDD based set bounds solvers the last form would typically be five separate constraints. The channelling component $j \in v \Leftrightarrow i \in v'$ is not explicitly represented. Instead, the underlying Boolean variables are re-used.

In Table 6, we use the model and search strategy used by Azevedo (2007), restricting the number of times a given element can occur in the sets $s_1, \ldots, s_m$. We compare against Gecode and Eclipse using the same MiniZinc model, as well as the published results of Cardinal and Length-Lex. The model used by the hybrid solver constructs one constraint for each pair of set variables, conjoining cardinality, intersection and ordering constraints. On those instances where a significant amount of search occurs, we again see a massive improvement beyond the performance of any of the pure set bounds propagation solvers. Our hybrid solver and Length-Lex are the most robust. We can see the hybrid requires the least search and is somewhat faster than Length-Lex. We also compare versus VSIDS search. The Steiner problems illustrate how a specialized search strategy can be better than the generic VSIDS approach. We can see that the Tseitin decomposition is not competitive for these problems

## 6.3 Fixed-weight Hamming Codes

The problem of finding maximal Hamming codes can also be expressed as a set-constraint problem. A Hamming code with distance $d$ and length $l$ is a set of $l$-bit codewords such that each pair of codewords must have at least $d$ bits which differ. A variation of this problem is to find maximal codes where all codewords have exactly $w$ bits set.

| Problem | Fixed-weight Hamming Codes | | | | | | | |
|---------|------|------|------|------|------|------|------|------|
| | Gecode | | Length-lex[‡] | | Static Hybrid | | | |
| | time | fails | time | fails | +f | +fs | +fi | fails |
| 8,4,4 | — | — | **0.07** | 110 | 0.16 | 0.17 | 0.16 | 897 |
| 9,4,3 | — | — | 2.05 | 4617 | 7.13 | 7.47 | 7.25 | 29985 |
| 9,4,4 | — | — | — | — | — | — | — | — |
| 9,4,5 | — | — | — | — | — | — | — | — |
| 9,4,6 | — | — | 0.40 | 908 | 1.67 | 1.66 | 1.61 | 10541 |
| 10,4,3 | — | — | 359.30 | 629822 | — | — | — | — |
| 10,4,4 | — | — | — | — | — | — | — | — |
| 10,4,5 | — | — | — | — | — | — | — | — |
| 10,4,6 | — | — | — | — | — | — | — | — |
| 10,4,7 | — | — | **1.99** | 4415 | — | — | — | — |
| 10,6,5 | 280.97 | 2175542 | **0.03** | 158 | 78.99 | 78.14 | 78.92 | 92349 |

Table 7: Results on hard Hamming instances with a static least-element in set search order, with no additional symmetry breaking.

| Problem | Fixed-weight Hamming Codes | | | | | |
|---------|------|------|------|------|------|------|
| | VSIDS Hybrid | | | | Tseitin | |
| | +f | +fs | +fi | fails | time | fails |
| 8,4,4 | 0.08 | 0.08 | 0.10 | 282 | 11.15 | 50530 |
| 9,4,3 | 0.30 | 0.26 | **0.24** | 1627 | 16.18 | 67876 |
| 9,4,4 | 56.55 | 45.95 | **43.39** | 210183 | — | — |
| 9,4,5 | 69.28 | 59.56 | **56.65** | 307786 | — | — |
| 9,4,6 | 0.43 | 0.39 | **0.36** | 2589 | 14.84 | 55292 |
| 10,4,3 | 102.64 | 90.37 | **86.09** | 638214 | — | — |
| 10,4,4 | 53.72 | 38.72 | **37.04** | 91781 | — | — |
| 10,4,5 | — | — | — | — | — | — |
| 10,4,6 | 509.13 | 404.65 | **385.48** | 987682 | — | — |
| 10,4,7 | 110.65 | 101.37 | 103.20 | 727465 | — | — |
| 10,6,5 | 0.71 | 0.57 | 0.54 | 5057 | 157.86 | 693148 |

Table 8: Results on hard Hamming instances with a VSIDS, with no additional symmetry breaking.

A formulation for this problem is:

$$\bigwedge_{i=1}^{m} (|s_i| = w) \ \wedge$$

$$\bigwedge_{i=1}^{m-1} \bigwedge_{j=i+1}^{m} (|s_i \oplus s_j| \geq d \wedge s_i < s_j)$$

where $s \oplus s' = (s - s') \cup (s' - s)$ is the symmetric difference. This is similar in structure to the formulation for the Steiner Systems; however, rather than having a fixed number of sets, we find the maximal code by repeatedly adding new sets and the corresponding constraints until no solution can be found. The unsatisfiability of $n$ codewords proves that the maximal code has $n-1$ codewords. We create BDD propagators for the constraint form $|v \oplus v'| \geq d \wedge v < v' \wedge |v| = w \wedge |v'| = w$.

We compare on two different models of the fixed-weight Hamming code problems, one just using the description above, and another where the first two sets are fixed to remove symmetries. We compare against Gecode, the published results of Length-Lex with our hybrid using a static search strategy (the same least element in set strategy as used for Social Golfers), as well as the hybrid solver and a Tseitin decomposition using VSIDS search. For our systems we compare with and without shortcutting and our optimized implementation. Since we are not sure which model was used by Length-Lex we report it results for both models.

| Problem | Fixed-weight Hamming Codes | | | | | | | |
|---------|------|------|------|------|------|------|------|------|
| | Gecode | | Length-lex[‡] | | Static Hybrid | | | |
| | time | fails | time | fails | +f | +fs | +fi | fails |
| 8,4,4 | 15.29 | 29869 | 0.07 | 110 | 0.04 | 0.04 | 0.04 | 51 |
| 9,4,3 | 66.72 | 216598 | 2.05 | 4617 | 0.28 | 0.29 | 0.28 | 2130 |
| 9,4,4 | — | — | — | — | 18.09 | 17.95 | 17.99 | 43318 |
| 9,4,5 | — | — | — | — | 55.90 | 56.52 | 57.14 | 71777 |
| 9,4,6 | 47.72 | 101832 | 0.40 | 908 | **0.04** | **0.04** | **0.04** | 208 |
| 10,4,3 | — | — | 359.30 | 629822 | — | — | — | — |
| 10,4,4 | — | — | — | — | — | — | — | — |
| 10,4,5 | — | — | — | — | — | — | — | — |
| 10,4,6 | — | — | — | — | — | — | — | — |
| 10,4,7 | — | — | 1.99 | 4415 | 6.16 | 6.24 | 6.29 | 22857 |
| 10,6,5 | 0.07 | 546 | 0.03 | 158 | **0.02** | **0.02** | **0.02** | 70 |

Table 9: Results on hard Hamming instances with a sequential least-element in set search order, with fixed first and second sets.

| Problem | Fixed-weight Hamming Codes | | | | | |
|---------|------|------|------|------|------|------|
| | VSIDS Hybrid | | | | Tseitin | |
| | +f | +fs | +fi | fails | time | fails |
| 8,4,4 | **0.03** | 0.05 | **0.03** | 61 | 1.06 | 6194 |
| 9,4,3 | 0.08 | **0.06** | **0.06** | 300 | 3.19 | 19952 |
| 9,4,4 | 1.28 | 1.10 | **1.04** | 4466 | 319.05 | 407762 |
| 9,4,5 | 4.82 | 4.20 | **4.03** | 21651 | 186.64 | 244474 |
| 9,4,6 | 0.06 | 0.05 | 0.06 | 256 | 1.31 | 8328 |
| 10,4,3 | 2.76 | 2.56 | **2.37** | 16755 | 120.22 | 226380 |
| 10,4,4 | 20.53 | 15.45 | **14.66** | 34503 | — | — |
| 10,4,5 | 143.50 | **104.29** | 104.39 | 184051 | — | — |
| 10,4,6 | 64.10 | 51.76 | **48.96** | 131379 | — | — |
| 10,4,7 | 2.21 | 2.05 | **1.96** | 13533 | 58.06 | 112269 |
| 10,6,5 | 0.03 | 0.04 | 0.03 | 145 | 0.10 | 1044 |

Table 10: Results on hard Hamming instances with a VSIDS search strategy, with fixed first and second sets.

Tables 7 to 10 show the results on the 11 hard instances reported by Hawkins et al. (2005). Clearly on these problems the VSIDS hybrid is the most robust. It can solve all but one instance in the basic model, and all with the additional symmetry breaking. This example also clearly shows the potential advantages of shortcutting and our improved data structures: these do not change the search but improve the time by 18% and 21% respectively for the base model, and 24% and 26% respectively for the improved model. Once more Tseitin decomposition is not competitive.

## 7. Related Work

Set-constraint problems have been an active area of research in the past decade. Many of the earlier solvers, beginning with PECOS (Puget, 1992), used the set-bounds repre-sentation combined with a fixed set of propagation rules for each constraint. This general approach was also used by Conjunto (Gervet, 1997), ECL$^i$PS$^e$ (IC-PARC, 2003), ILOG Solver (ILOG, 2004) and Mozart (Müller, 2001). However, as set-bounds are a relatively weak approximation of the domain of a set variable, a variety of variations have been developed to improve the propagation strength of set-constraint solvers. These include solvers which combine set-bounds representation with either cardinality information, such as that proposed by Azevedo (2002, 2007), lexicographic bounds information (Sadler & Gervet, 2004) or both (Gervet & Van Hentenryck, 2006; Yip & Van Hentenryck, 2009).

BDD-based approaches to set-constraint solving, such as that presented by Hawkins et al. (2005) differs greatly from these approaches, as it is possible to perform propagation over arbitrary constraints; Lagoon and Stuckey (2004) also demonstrated the feasibility of a BDD-based solver which maintains a complete domain representation of set variables.

These directly BDD-based algorithms were used to construct the earlier hybrid solver presented by Hawkins and Stuckey (2006), which is conceptually similar to the solver presented in this paper. The solver presented here is much more efficient, and includes improvements such as filtering and shortcutting not present in the solver of Hawkins and Stuckey (2006). The solver of Damiano and Kukula (2003) also combines BDD solving and SAT solving, but rather than building BDDs from a high-level problem description and lazily constructing a SAT representation, instead takes a CNF SAT representation and constructs a BDD from a collection of clauses with the primary goal of variable elimination. It is essentially equivalent to the `base` solver.

The underlying BDD propagation algorithm is similar to propagation of the `case` constraint of SICStus PRolog (SICS, 2009) and Multi-valued Decision Diagrams (MDDs) (see e.g., Cheng & Yap, 2008). Indeed we have adapted the dead subgraph memoization and shortcutting devices of Cheng and Yap (2008) to BDD propagation. Propagators for `case` and MDDs do not presently use filtering or generate reasons.

Finally the hybrid set solver we present in this paper is an example of a *lazy clause generation solver* (Ohrimenko, Stuckey, & Codish, 2007, 2009). The BDD propagators can be understood as lazily creating a clausal representation of the set constraints encoded in the BDD, as search progresses.

## 8. Concluding Remarks

In this paper we have improved BDD-based techniques for set-bounds propagation, having demonstrated an approach which avoids the need for expensive BDD construction and manipulation operations. This traversal-based method, when combined with filtering to reduce the number of redundant propagator executions and dead subgraph memoization and shortcutting, is at least an order of magnitude faster than previous techniques which construct BDDs during runtime (Hawkins et al., 2005).

Furthermore, when integrated into a modern SAT solver with clause learning and augmented with a method for generating nogoods, the new hybrid solver is capable of solving hard problem instances several orders of magnitude faster than pure bounds set solvers. Overall the hybrid solver is robust and highly competitive with any other propagation based set-solvers we are aware of.

In many set problems there are significant numbers of symmetries and there is a large body of work solving set problems with symmetry breaking techniques (see e.g., Puget, 2005). It would be interesting to combine symmetry breaking with our hybrid solver.

## 9. Acknowledgments

Part of this work was published previously (Gange, Lagoon, & Stuckey, 2008). NICTA is funded by the Australian Government as represented by the Department of Broadband, Communications and the Digital Economy and the Australian Research Council.

## References

Azevedo, F. (2002). *Constraint Solving over Multi-valued Logics*. Ph.D. thesis, Faculdade de Ciências e Tecnologia, Universidade Nova de Lisboa.

Azevedo, F. (2007). Cardinal: A finite sets constraint solver. *Constraints*, *12*(1), 93–129.

Bryant, R. (1986). Graph-based algorithms for Boolean function manipulation. *IEEE Trans. Comput.*, *35*(8), 677–691.

Cheng, K., & Yap, R. (2008). Maintaining generalized arc consistency on ad hoc r-ary constraints. In *14th International Conference on Principles and Process of Constraint Programming*, pp. 509–523.

Damiano, R., & Kukula, J. (2003). Checking satisfiability of a conjunction of BDDs. In *Proceedings of Design Automation Conference*, pp. 818–823.

Davis, M., Logemann, G., & Loveland, D. (1962). A machine program for theorem-proving. *Communications of the ACM*, *5*, 394–397.

Eén, N., & Sörensson, N. (2003). An extensible SAT-solver. In Giunchiglia, E., & Tacchella, A. (Eds.), *Proceedings of SAT 2003*, Vol. 2919 of *LNCS*, pp. 502–518.

Eén, N., & Sörensson, N. (2006). Translating pseudo-boolean constraints into SAT. *Journal on Satisfiability, Boolean Modeling and Computation*, *2*, 1–26.

Gange, G., Lagoon, V., & Stuckey, P. (2008). Fast set bounds propagation using BDDs. In *18th European Conference on Artificial Intelligence*, pp. 505–509.

GECODE (2008). Gecode. www.gecode.org. Accessed Jan 2008.

Gervet, C. (1997). Interval propagation to reason about sets: Definition and implementation of a practical language. *Constraints*, *1*(3), 191–246.

Gervet, C., & Van Hentenryck, P. (2006). Length-lex ordering for set CSPs. In *Proceedings of the National Conference on Artificial Intelligence*, pp. 48–53.

Hawkins, P., Lagoon, V., & Stuckey, P. (2004). Set bounds and (split) set domain propagation using ROBDDs. In *17th Australian Joint Conference on Artificial Intelligence*, Vol. 3339 of *LNCS*, pp. 706–717.

Hawkins, P., Lagoon, V., & Stuckey, P. (2005). Solving set constraint satisfaction problems using ROBDDs. *Journal of Artificial Intelligence Research*, *24*, 106–156.

Hawkins, P., & Stuckey, P. (2006). A hybrid BDD and SAT finite domain constraint solver. In *Proceedings of the 8th International Symposium on Practical Aspects of Declarative Languages*, Vol. 3819 of *LNCS*, pp. 103–117.

IC-PARC (2003). The ECLiPSe constraint logic programming system. [Online, accessed Oct 2008]. `http://www.eclipse-clp.org/`.

ILOG (2004). ILOG Solver. [Online, accessed Oct 2008]. `http://www.ilog.com/`.

Lagoon, V., & Stuckey, P. (2004). Set domain propagation using ROBDDs. In *Proceedings of the 10th International Conference on Principles and Practice of Constraint Programming*, Vol. 3258 of *LNCS*, pp. 347–361.

van Lint, J. H., & Wilson, R. M. (2001). *A Course in Combinatorics* (2nd edition). Cambridge University Press.

Müller, T. (2001). *Constraint Propagation in Mozart*. Doctoral dissertation, Universität des Saarlandes, Naturwissenschaftlich-Technische Fakultät I, Fachrichtung Informatik, Saarbrücken, Germany.

Ohrimenko, O., Stuckey, P., & Codish, M. (2007). Propagation = lazy clause generation. In Bessiere, C. (Ed.), *Proceedings of the 13th International Conference on Principles and Practice of Constraint Programming*, Vol. 4741 of *LNCS*, pp. 544–558. Springer-Verlag.

Ohrimenko, O., Stuckey, P., & Codish, M. (2009). Propagation via lazy clause generation. *Constraints*, *14*(3), 357–391.

Puget, J.-F. (1992). PECOS: a high level constraint programming language. In *Proceedings of SPICIS'92*, Singapore.

Puget, J.-F. (2005). Symmetry breaking revisited. *Constraints*, *10*(1), 23–46.

Sadler, A., & Gervet, C. (2004). Hybrid set domains to strengthen constraint propagation and reduce symmetries. In Wallace, M. (Ed.), *Proceedings of the 10th International Conference on Principles and Practice of Constraint Programming (CP04)*, No. 3258 in LNCS. Springer-Verlag.

SICS (2009). Sicstus prolog. www.sics.se/sicstus.

Subbarayan, S. (2008). Efficent reasoning for nogoods in constraint solvers with BDDs. In *Proceedings of Tenth International Symposium on Practical Aspects of Declarative Languages*, Vol. 4902 of *LNCS*, pp. 53–57.

Tseitin, G. (1968). On the complexity of derivation in propositional calculus. *Studies in Constructive Mathematics and Mathematical Logic, Part 2*, 115–125.

Yip, J., & Van Hentenryck, P. (2009). Evaluation of length-lex set variables. In *Proceedings of the 15th International Conference on Principles and Practice of Constraint Programming*, pp. 817–832.